%% file: main.tex
\documentclass[preprint]{article}

\include{config/packages}
\include{config/math}

\include{config/acronyms}
\include{config/theorems}

\usepackage{styles/icml2026}

\icmltitlerunning{Scaling Laws for Uncertainty in Deep Learning}

\begin{document}

\twocolumn[
  \icmltitle{Scaling Laws for Uncertainty in Deep Learning}



  \icmlsetsymbol{equal}{*}

  \begin{icmlauthorlist}
    \icmlauthor{Mattia Rosso}{kaust}
    \icmlauthor{Simone Rossi}{eurecom}
    \icmlauthor{Giulio Franzese}{eurecom}
    \icmlauthor{Markus Heinonen}{aalto}
    \icmlauthor{Maurizio Filippone}{kaust}
  \end{icmlauthorlist}

  \icmlaffiliation{kaust}{King Abdullah University of Science and Technology (KAUST), Saudi Arabia}
  \icmlaffiliation{eurecom}{EURECOM, France}
  \icmlaffiliation{aalto}{Aalto University, Finland}

  \icmlcorrespondingauthor{Mattia Rosso}{mattia.rosso@kaust.edu.sa}

  \icmlkeywords{Machine Learning, ICML}

  \vskip 0.3in
]



\printAffiliationsAndNotice{}  % no special notice (required even if empty)

\begin{abstract}
    Scaling laws in deep learning describe the predictable relationship between a model's performance, usually measured by test loss, and key design choices such as dataset and model size. Inspired by these findings, we investigate a parallel direction: do similar scaling laws govern predictive uncertainties in deep learning? 
    Identifiable models under a Bayesian framework exhibit $O(1/N)$ epistemic uncertainty contraction rate, with respect to dataset size $N$. However, these guarantees fail in over-parameterized regimes, where scaling behaviors remain largely unexplored.
    Across vision and language tasks, we show that in- and out-of-distribution predictive uncertainties, estimated via popular Bayesian and ensemble methods, follow predictable scaling laws relative to model and dataset size. 
    Besides the elegance of scaling laws and the practical utility of extrapolating uncertainties to larger data or models, this work provides strong evidence to dispel recurring skepticism against Bayesian approaches: \textit{``In many applications of deep learning we have so much data available: what do we need Bayes for?''} Our findings show that \textit{``so much data''} is typically not enough to make epistemic uncertainty negligible.
\end{abstract}

\section{Introduction}

\begin{figure}
    \centering
    \includegraphics[width=0.45\textwidth]{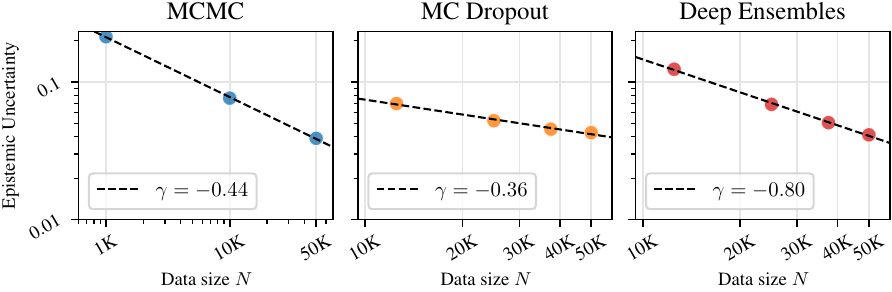}
    \caption{\textbf{Deep learning uncertainty is predictable, empirically}. ResNet-18 Epistemic Uncertainty scaling with the number of training data $N$ on \texttt{CIFAR-10}.}
    \label{fig:first_page_plot}
\end{figure}

Empirical \textit{scaling laws} describe how deep learning performance improves predictably with increasing data, parameters, or compute, typically following a power-law decay

\begin{equation}
    f(x) \propto x^\gamma, \quad \gamma < 0
\end{equation}

where $x \in \{N,P,C\}$ denotes dataset size, parameter count, or compute budge \citep{Kaplan2020,Hoffmann2022}. A related `double descent` effect shows that increasing model capacity can surprisingly improve generalization \citep{Belkin2019}. These findings suggest that bigger models often perform better. A parallel question is: do similar scaling laws govern uncertainty in deep learning?

\paragraph{Bayesian Deep Learning}
Bayesian deep learning provides a principled framework for quantifying uncertainty in neural networks by marginalizing weights, rather than relying on point estimates, to obtain predictive uncertainties \citep{Neal1996,MacKay1995}.
Recent contributions show that, under rather simple conditions, inferring only a subset of model parameters can retain the capacity to represent any predictive distribution, at a significantly lower cost \citep{Sharma2023}. 
The ability to produce well‐calibrated predictive distributions and uncertainties is essential in decision‐making and safety‐critical applications, such as medical diagnosis and autonomous driving, where sound quantification of uncertainty can guide downstream choices \citep{Papamarkou2024}.
% Applications on active learning?
%
However, exact Bayesian inference in large networks is intractable, which has motivated scalable approximations such as those based on the Laplace method \citep{Ritter2018, Antoran2022}, \gls{VI} \citep{Graves2011, Shen2024}, \gls{MCD} \citep{Gal2016}, along with (non-Bayesian) \gls{DE} \citep{Lakshminarayanan2017}, which have proven effective and practical in obtaining sensible predictive uncertainties. Sampling-based methods, such as \gls{SGLD} \citep{Welling2011} and \gls{SGHMC} \citep{Chen2014} provide scalable \gls{MCMC} techniques for posterior inference over neural network weights. In the remainder of this paper, we will generally refer to \gls{UQ} methods to indicate Bayesian and non-Bayesian approaches designed to obtain predictive uncertainties.

\paragraph{Predictive Uncertainty and Scaling}
Despite recent criticism \citep{Wimmer2023}, current trends have focused on information-theoretic decomposition of predictive uncertainties into aleatoric and epistemic parts \citep{Hullermeier2021}.
In this work, we study how \emph{predictive uncertainties} evolve in deep models as dataset size $N$ and model size $P$ grow, evaluating a range of \gls{UQ} methods and architectures for both vision and language modalities. With stronger emphasis on scaling with respect to $N$, we  systematically vary optimization settings, prior choices, stochasticity level, and inference techniques, demonstrating that different configurations exhibit a range of scaling behaviors across uncertainty metrics, but crucially all of them follow power-laws (see, e.g., \cref{fig:first_page_plot}).

While classic results show $O(1/N)$ posterior-variance contraction in identifiable parametric models \citep{Cam1953}, over-parameterized neural networks present challenges due to singular likelihoods, non-identifiability, and non-Gaussian posteriors. We specifically study the \gls{LLA} in \cref{sec:lla} and confirm that, under network linearization, the posterior variance scaling asymptotically approaches the $1/N$ trend.

Although empirical scaling laws apply to test performance \citep{Kaplan2020}, a systematic understanding of how uncertainty scales with $N$, model size $P$, or computational budget is lacking. Both Bayesian and non-Bayesian \gls{UQ}  methods yield ensembles, with \gls{EU} reflecting prediction diversity. One of the interesting aspects of scaling laws for uncertainty is to identify when \gls{EU} vanishes, indicating ensemble collapse.

\paragraph{\gls{SLT}}
% Several theories characterize neural networks in the large-data limit. \gls{SLT} models deep networks as singular statistical manifolds, claiming that singularities of loss minima determine learning behavior, and that generalization follows a power-law $\mathcal{L}-\mathcal{L}_0 = \frac{c}{n^\lambda}$ \citep{Watanabe2009}. In this expression, $\mathcal{L}$ indicates the test loss, $\mathcal{L}_0$ the theoretical minimal achievable loss (e.g., due to noise in the data), $c$ a constant that depends on the model and $\lambda$ the \gls{RLCT} or learning coefficient. This last quantity governs the rate at which the generalization loss approaches the irreducible loss as $n$ (number of training data points) grows. 
Several theories characterize neural networks in the large-data limit. \gls{SLT} models deep networks as singular statistical manifolds and shows that singularities of loss minima determine learning behavior. In particular, for Bayesian learning it predicts that the expected generalization error $G_n$ (the expected excess loss when going from $n$ to $n{+}1$ training samples) decays as
$\E[G_n] = \frac{\lambda}{n} + o\!\left(\frac{1}{n}\right)$, where $\lambda$ is the \gls{RLCT} (learning coefficient) \citep{Watanabe2009}. This last quantity governs the rate at which the generalization loss approaches the irreducible loss as $n$ (number of training data points) grows. We derive a formal connection between the generalization error and the \gls{TU} in the case of linear models. We believe that such a framework may help explaining the scaling behavior of uncertainty in deep models. 

Other perspectives include the manifold dimensionality hypothesis, linking generalization to the intrinsic dimensionality of learned representations \citep{Ansuini2019}, and mechanistic interpretability efforts explaining simple scaling phenomena via emergent internal circuits \citep{Nanda2023}.

\paragraph{Contributions}
In summary, our contributions are as follows:
 
  (i) \emph{Empirical study.}  We provide a comprehensive evaluation of predictive uncertainties using a variety of \gls{UQ} methods across different architectures, modalities, and datasets. To the best of our knowledge, this is the first study to consider scaling laws associated with any form of uncertainty in deep learning.
  
  (ii) \emph{Scaling patterns.}  We empirically demonstrate that predictive uncertainties evaluated on in- and out-of-distribution, follow power-law trends with dataset and model size. This allows us to extrapolate to large dataset sizes and to identify data regimes where \gls{UQ} approaches remain relevant to characterize the diversity of the ensemble to a given numerical precision. 

 (iii) \emph{Theoretical insights.}  We derive a formal connection between generalization error in \gls{SLT} and \gls{TU} in linear models. This novel analysis provides an interesting lead to explain the scaling laws observed in the experiments for over-parameterized models.

\section{Background}

\glsreset{EU}
\subsection{Uncertainty Quantification}
\label{sec:uq}

Predictive uncertainty refers to metrics associated with an ensemble of predictive distributions and it can be decomposed into \gls{AU}, arising from intrinsic data variability, and \gls{EU}, reflecting uncertainty due to limited data or model knowledge \citep{Hullermeier2021}.

The \gls{TU} is the entropy of the mean predictive distribution:

\begin{align}
    \mathrm{TU}(\mbx) &= \mathbb{H}\left[\frac{1}{K}\sum_{k=1}^K p\Big(y | \mbx, \mbtheta^{(k)}\Big)\right],
\end{align}

where $\mbtheta^{(k)}$ are the model parameters of the $k$'th ensemble member or stochastic pass, and $p(y | \mbx, \mbtheta^{(k)})$ is its predictive distribution. The irreducible uncertainty \gls{AU} is the average entropy of predictions:

\begin{align}
    \mathrm{AU}(\mbx) &= \frac{1}{K}\sum_{k=1}^K \mathbb{H}\left[ p\Big(y|\mbx,\mbtheta^{(k)}\Big) \right].
\end{align}

Finally, the reducible uncertainty \gls{EU} is their difference,

\begin{align}
    \mathrm{EU}(\mbx) &= \mathrm{TU}(\mbx) - \mathrm{AU}(\mbx).
\end{align}

In this work, we study the power-law of test predictive uncertainty, averaged over all the (fixed) test samples:

\begin{align}
    \frac{1}{N_{\mathrm{test}}}\sum_{n=1}^{N_{\mathrm{test}}} \mathrm{U}(\mbx_n), \qquad \mathrm{U} \in \{\mathrm{TU}, \mathrm{AU}, \mathrm{EU}\}.
\end{align}

\paragraph{Limitations of \gls{AU} and \gls{EU}}
While popular, these entropy-based metrics have limitations:
(i) The standard decomposition of \gls{TU} assumes additive separation of \gls{AU} and \gls{EU}, which \citet{Wimmer2023} argues not to hold in complex deep models, and the difficulty of disentangling them hinders their interpretability \citep{DeJong2025};
(ii) Recent works highlight \gls{EU} collapse of large ensembles, leading to overly confident predictions \citep{Kirsch2024, Fellaji2024}.

Despite these criticisms, these metrics remain useful and tractable in practice, especially when paired with typical \gls{UQ} methods. Their consistency across tasks such as active learning, out-of-distribution detection and model calibration makes them valuable diagnostic tools.

\subsection{Scaling Laws and Generalization}

Empirical studies in deep learning have demonstrated that performance metrics scale predictably with model size, dataset size and compute.
Initial findings by \citet{Hestness2017} and \citet{Kaplan2020} showed that test loss typically decreases following a power-law of the form 

\begin{equation}
    \mathcal{L}(x) = \left( \frac{x_0}{x} \right)^{\alpha} + \mathrm{const.}\,,
\end{equation}

% \begin{equation}
%     \mathcal{L}(x) = \mathcal{L}_\infty + \left( \frac{x_0}{x} \right)^{\alpha}\,,
% \end{equation}

where $x$ is the resource under analysis (e.g., dataset size $N$, model size $P$, or compute budget $C$) 
% $\mathcal{L}_\infty$ is the irreducible loss, 
and $\alpha$ is modality- and task-specific ($x_0$ is a reference constant).
These relationships hold consistently across model architectures, tasks, and modalities, as empirically shown by \citet{Henighan2020}, who demonstrated similar loss scaling for language, image, video and multimodal domains.
The scaling exponents differ across domains, but the functional form of the scaling law remains stable, indicating a general underlying behavior.

Theoretical explanations of scaling behaviors are only recently emerging: for example \citet{sharma2022scaling,bahri2024explaining} linked $\alpha$ to the intrinsic dimension and spectral properties of the data, showing how the data geometry drives the reducible portion of the loss.
However, despite these empirical and theoretical advances, a comprehensive understanding of how \gls{EU} and \gls{AU} scale with model and data size remain an open question. 
Indeed, a Bayesian perspective on uncertainty scaling laws is currently missing: it is unclear what kind of power laws uncertainty exhibits, if any. Understanding uncertainty scaling could significantly enhance the development and comprehension of Bayesian deep learning. 

\section{Methods}

In this section, we describe the approximate Bayesian inference and ensemble methods used in our experiments. We hypothesize that if uncertainty scaling laws exist, they should emerge regardless of the \gls{UQ} method.

\paragraph{Monte Carlo Dropout.}

\gls{MCD} is a simple and common inference method, with connections to \gls{VI} \citep{Gal2016}. During training, standard dropout is applied, while at test time dropout masks are resampled to produce stochastic forward passes, yielding ensemble predictions. Due to its simplicity and universality, \gls{MCD} is a good \gls{UQ} baseline. 

\paragraph{Gaussian Approximations.}

Gaussian are a common family of posterior approximations, encompassing both Laplace methods \citep{Ritter2018} and variational inference (\gls{VI}) \citep{Graves2011}.
While Laplace relies on the local curvature of the log-posterior to define the Gaussian covariance, \gls{VI} instead directly optimizes an approximate distribution to match the true posterior in the \gls{KL} sense. Building on this line of work, we also include experiments with the \gls{IVON} optimizer \citep{Shen2024}, which unifies natural-gradient \gls{VI} \citep{Khan2023} with an online Newton method to achieve efficient large-scale Bayesian learning.

\paragraph{Markov Chain Monte Carlo.}

\gls{MCMC} is the classic method of obtaining samples from the posterior over model parameters \citep{Neal1996,MacKay1995}. Gradient-based \gls{MCMC}, such as \gls{HMC}, are some of the most effective samplers. Mini-batch-based \gls{SGHMC} \citep{Chen2014} and Langevin dynamics \citep{Welling2011} have been successfully proposed to sample from the posterior over parameters of deep neural networks of moderate size \citep{Tran2022,izmailov2021bayesian}. 
We consider parameter sampling with both global parameter priors and Gaussian process functional priors \citep{Tran2022}.

\paragraph{Deep Ensembles.}

\gls{DE} \citep{Lakshminarayanan2017} are a popular technique to obtain predictive uncertainties, despite lacking a fully Bayesian interpretation. Multiple networks are trained independently from different seeds and uncertainty is estimated from the ensemble of predictions. \gls{DE} tend to provide better uncertainty estimates and are more robust to model misspecification than \gls{MCD}. We train ensembles of size $M \in \{5,10\}$.

\paragraph{Partially stochastic networks.}
We consider partially stochastic networks where only few layers are inferred, while the rest are optimized in a standard way \citep{Sharma2023}.

\section{Experiments}

In this section, we provide a detailed analysis of our experimental results. We first highlight the most significant findings in the vision domain, focusing on the scaling behavior of uncertainty both in-distribution and \gls{OOD} (\cref{sec:image-domain}). This is complemented by an analysis of uncertainty extrapolation, detailed in \cref{app:fig:wideresnets-cifar10-extrapolation}. We report an experiment in the text classification setting using a GPT-2 language model (\cref{sec:text-domain}). For completeness, we also include a toy experiment with a Bayesian \gls{MLP} (\cref{sec:mlp-extrapolation}). Our study covers a wide range of configurations across architectures, datasets, and \gls{UQ} setups, providing a comprehensive view of scaling trends.

\subsection{Image classification}
\label{sec:image-domain}

We use common image classification architectures of ResNet \citep{He2015}, WideResNet \citep{Zagoruyko2016} and Vision Transformer (ViT) \citep{Dosovitskiy2021}. We report the \texttt{CIFAR-10} results and part of the \texttt{ImageNet32} and \texttt{CIFAR-100} results in the main paper, while additional analyses are provided in \cref{sec:app:additional_results}. 

\subsubsection{ResNet and WideResNet}

If not specified, we train these models for $400$ epochs with \gls{SGD} optimizer (momentum $0.9$ and weight decay $5\times10^{-4}$); in some experiments we adopt a cosine annealing scheduler on the learning rate and in others we do not. See \cref{sec:app:experimental_setup} for more experimental details.

\paragraph{Monte Carlo Dropout} 

\begin{figure}[t]
    \centering
    \includegraphics[width=0.45\textwidth]{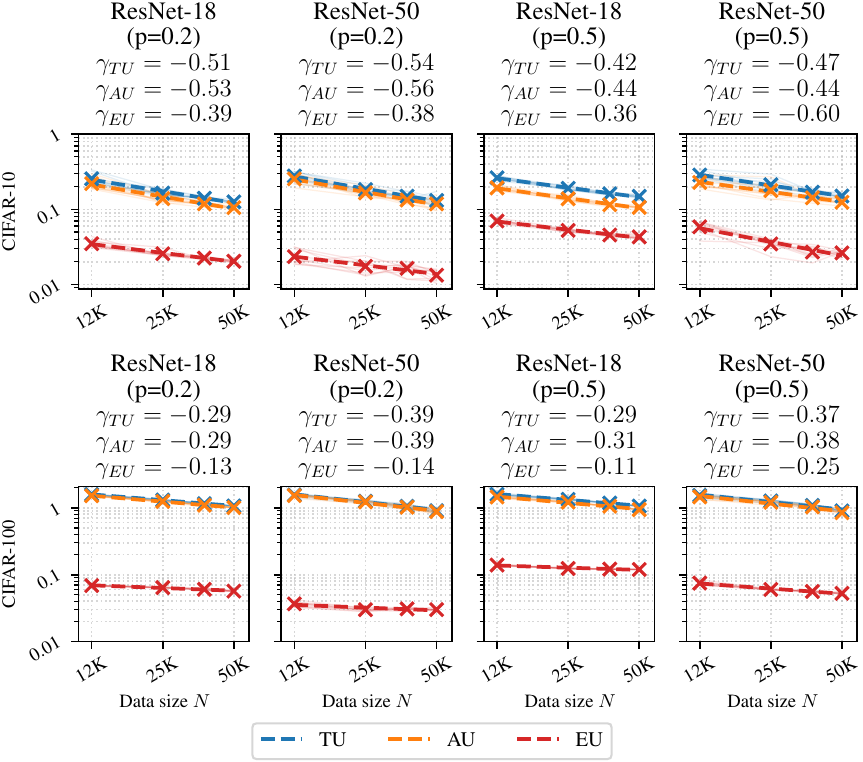}
    \caption{\textbf{ResNet-$\mathbf{d}$  uncertainty scaling on \texttt{CIFAR-10} and \texttt{CIFAR-100} datasets}: We use \gls{MCD} (with fixed dropout rate $p=0.2$ and $p=0.5$); each point $\times$ corresponds to the average over $10$ independent folds (varying both data subsampling and model initialization). We consider $25\%$, $50\%$, $75\%$ and $100\%$ subsets of the training dataset. Dashed lines represent linear regressions fitted to the mean uncertainty metrics (\gls{AU}, \gls{EU}, \gls{TU}) on a fixed test set (see \cref{sec:uq}), following a power-law decay of the form $N^{\gamma_{TU}}$, $N^{\gamma_{AU}}$, and $N^{\gamma_{EU}}$. Both axes are shown on a logarithmic scale.}
    \label{fig:resnet-cifar10-cifar100}
\end{figure}

\begin{figure}[t]
  \centering
     \includegraphics[width=0.45\textwidth]{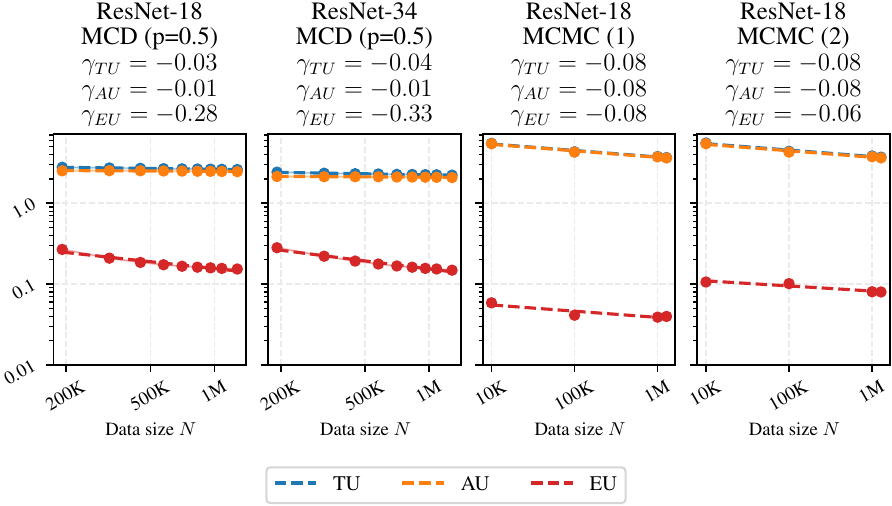}
  \caption{\textbf{Uncertainty estimates of ResNets on \texttt{ImageNet-32} dataset}: The first two subplots report \gls{MCD} ($p=0.5$) uncertainties for ResNet-18 and ResNet-34, trained with fixed learning rate $10^{-3}$ on $9$ increasing subsets of the training data. We also report uncertainties obtained through \gls{MCMC} considering only the first layer stochastic (\gls{MCMC} (1)) and only the first two layers stochastic (\gls{MCMC} (2)) for ResNet-18 trained on $4$ increasing subsets of the training data.}
  \label{fig:resnet_imagenet32_mcd_mcmc_scaling}
\end{figure}

In ResNets, we add dropout layers after the convolutional blocks and before the fully connected layers following \citet{Kim2023}. For WideResNets, we adopt the official implementation by \citet{Zagoruyko2016}, where dropout is applied between the two convolutional layers within each residual unit. We experiment with various dropout rates finding more expressivity in the scaling laws obtained with $p=0.5$, especially in ResNets architectures \citep{Gal2017}. Some results are reported in \cref{fig:first_page_plot}, \cref{fig:resnet-cifar10-cifar100}, and \cref{fig:resnet_imagenet32_mcd_mcmc_scaling}. 

Across our experiments, we observe that the \gls{EU} is typically smaller than the \gls{AU}. Regarding the decreasing trend of \gls{AU} that we observe across various experiments, \citet{Wimmer2023} also report that \gls{AU} can decrease under limited data, making estimates potentially unreliable in such regimes. Prior work \citep{DeJong2025,Mucsanyi2024} further shows that \gls{AU} and \gls{EU} often correlate in image classification, suggesting entanglement in practice.

\paragraph{Monte Carlo Dropout combined with Sharpness Aware Minimization}

\begin{figure}[t]
    \centering
    \includegraphics[width=0.45\textwidth]{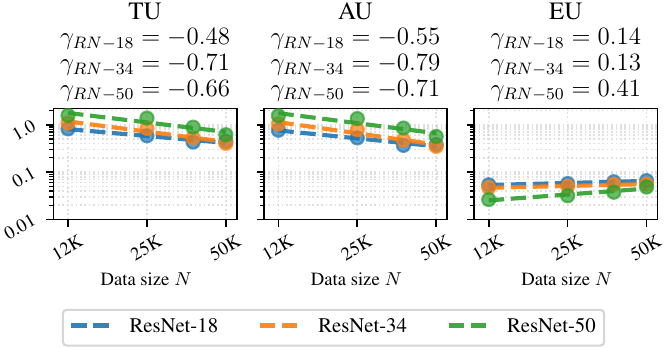}
    \caption{\textbf{Impact of \gls{SGD}+\gls{SAM} on uncertainty scaling}: ResNets on \texttt{CIFAR-10} dataset using \gls{MCD} ($p = 0.5$) for \gls{UQ}. \gls{SAM} biases solutions towards flatter minima and the combination with \gls{MCD} preserves functional diversity as data size increases.}
    \label{fig:cifar10-resnets+sam}  
\end{figure} 

We also study \gls{MCD} combined with \gls{SAM} \citep{Foret2020} to assess how two generalization methods interact in terms of uncertainty. We hypothesize that the increasing \gls{EU} in \cref{fig:cifar10-resnets+sam} arises because \gls{SAM}, by avoiding sharp minima, is forced into flatter basins as $N \to \infty$: by selecting flatter basins under increased curvature, the optimizer may avoid regions of low epistemic variance and instead favor broader regions spanning more diverse functions. This aligns with recent findings that \gls{SAM} can enhance ensemble diversity by trading off sharpness minimization with data-driven de-correlation \citep{Lu2024}, while \gls{MCD} further promotes functional diversity via implicit bagging.

\begin{figure}[t]
    \centering
    \includegraphics[width=0.45\textwidth]{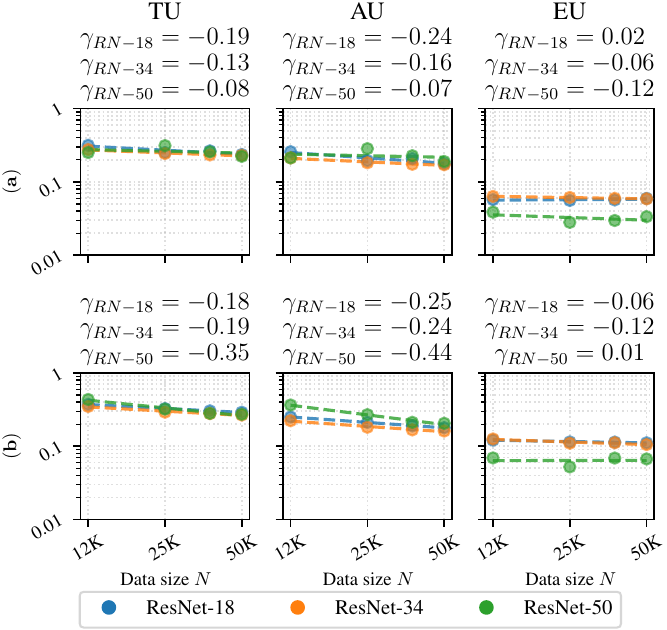}
    \caption{\textbf{ResNets uncertainties out-of-distribution:} We use \gls{MCD} with $p = 0.2$ in \textbf{(a)} and $p = 0.5$ in \textbf{(b)}. For models trained on incrementally larger training subsets of \texttt{CIFAR-10}, we report the predictive uncertainties when testing on the (whole) \texttt{CIFAR10-C} dataset, averaged over all corruption levels ($1$-$5$) and corruption types considered. \gls{OOD}, we expect to observe larger uncertainties - \gls{EU}, for instance, should decay gradually as the data space becomes increasingly populated with additional samples within the same domain (in-fill).}
    \label{fig:cifar10-ood}  
\end{figure} 

\paragraph{Uncertainty out-of-distribution}
We report results training on \texttt{CIFAR-10} and testing on \texttt{CIFAR-10-C} (corrupted), which allows us to explore how uncertainty behaves \gls{OOD}; results in \cref{fig:cifar10-ood}. 

\begin{figure}[tb]
    \centering
    \includegraphics[width=0.45\textwidth]{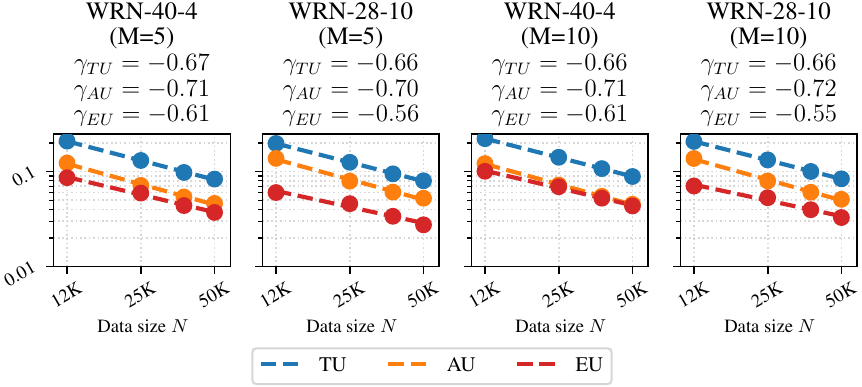}
    \caption{\textbf{WideResNet -$\mathbf{w}$-$\mathbf{d}$ uncertainty scaling on \texttt{CIFAR-10} dataset}: We consider WideResNet-40-4 and WideResNet-28-10 and perform \gls{UQ} using \gls{DE} with $M=5$ and $M=10$ ensemble members.}
    \label{fig:wideresnet-de}
\end{figure}

\paragraph{Deep Ensembles}
In \cref{fig:wideresnet-de}, we report scaling laws for \gls{DE} \citep{Lakshminarayanan2017}, using $M \in \{5, 10\}$ independently trained models. 
Even for this \gls{UQ} approach, we observe the emergence of power-law scalings of uncertainties with respect to the number of data.

\paragraph{Markov Chain Monte Carlo}
We also observe a similar behavior for \gls{MCMC}. 
In these experiments, we choose a weakly informative prior over all parameters with  zero mean and standard deviation of $10$. 
In \cref{sec:app:additional_results}, we also report results for \gls{MCMC} after optimizing the priors according to the approach in \cite{Tran2022}, where we target a \gls{GP} with isotropic covariance with log-length-scale of $\frac{1}{2}\log{D}$ and and log-marginal variance of $2$. 
In \cref{fig:resnet_imagenet32_mcd_mcmc_scaling}, \gls{MCMC}~(1) and \gls{MCMC}~(2) refer to the approach of treating the first and first+second layers of the model in a stochastic fashion, while the other parameters are kept fixed to a pretrained solution \citep{Sharma2023}. 

\paragraph{ResNets uncertainty scaling with model parameters}

\begin{figure}[t]
    \centering
    \includegraphics[width=0.45\textwidth]{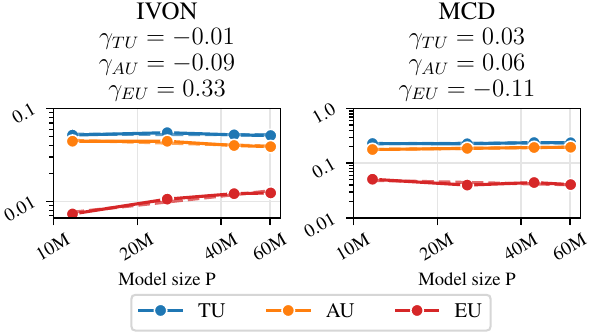}
    \caption{\textbf{Uncertainty scaling with model parameters of ResNets on \texttt{CIFAR-10}:} We evaluate a wide spectrum of ResNet architectures on \texttt{CIFAR-10} dataset. The corresponding parameter counts are: ResNet-18 ($11.7 M$), ResNet-50 ($25.6M$), ResNet-101 ($44.5M$), and ResNet-152 ($60.2M$). \textit{Left:} We train each model on the entire dataset using the \gls{IVON} optimizer for $200$ epochs, following the setup of \citet{Shen2024} (details in \cref{sec:app:experimental_setup}). \textit{Right:} We compare against \gls{MCD} ($p = 0.5$), training for $400$ epochs with \gls{SGD} and fixed learning rate $10^{-3}$.}
    \label{fig:ivon_mcd_scale_P}  
\end{figure}

\begin{figure}[!htbp]
    \centering
    \includegraphics[width=0.45\textwidth]{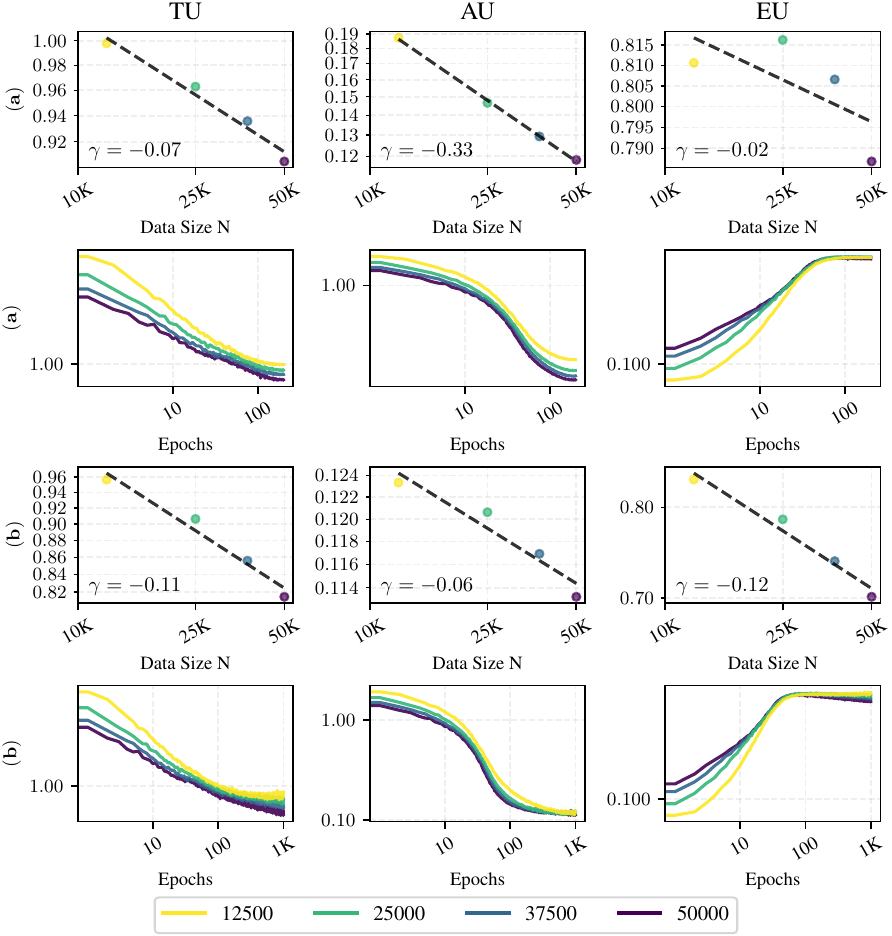}
    \caption{\textbf{ViT-small uncertainty training dynamics on \texttt{CIFAR-10} dataset:} In \textbf{(a)} we use \gls{MCD} (fixed $p=0.5$ both in the embeddings and in the transformer blocks) and we train the model for $200$ epochs with Adam optimizer \citep{Kingma2017} and cosine annealing. In \textbf{(b)} we train the same model for $1000$ epochs and fixed learning rate $10^{-4}$. The training dynamics show that the the shape/speed of convergence of of \gls{TU}, \gls{AU}, \gls{EU} depends on the optimization trajectories underneath.}
    \label{fig:cifar10-ViT-two-experiments-learning}  
\end{figure} 

We investigate uncertainty scaling with model size in a preliminary experiment using \gls{MCD} and the \gls{IVON} optimizer. \gls{IVON} shows the expected increase in \gls{EU} with larger models, whereas \gls{MCD} does not, likely reflecting limitations of the inference scheme. 
Parameter permutation symmetries (e.g., swapping hidden units or attention heads) generate exponentially many modes that yield the same function. Recent works \citep{Rossi2023, Laurent2023, Gelberg2024} show these modes are connected by low-loss paths and are not functionally distinct. Thus, increasing model capacity adds redundancy rather than distinct hypotheses, explaining the weak or flat \gls{EU} dependence observed in \cref{fig:ivon_mcd_scale_P}.

% \begin{figure}[t]
%     \centering
%     \includegraphics[width=0.45\textwidth]{figures/ICML/vit_schedulers.pdf}
%     \caption{\textbf{ViT-small uncertainty training dynamics on \texttt{CIFAR-10} dataset:} In \textbf{(a)} we use \gls{MCD} (fixed $p=0.5$ both in the embeddings and in the transformer blocks) and we train the model for $200$ epochs with Adam optimizer \citep{Kingma2017} and cosine annealing. In \textbf{(b)} we train the same model for $1000$ epochs and fixed learning rate $10^{-4}$. The training dynamics show that the the shape/speed of convergence of of \gls{TU}, \gls{AU}, \gls{EU} strongly depends on the optimization trajectories underneath.}
%     \label{fig:cifar10-ViT-two-experiments-learning}  
% \end{figure} 

\subsubsection{Vision Transformer}

% \begin{figure}[t]
%     \centering
%     \includegraphics[width=0.45\textwidth]{figures/ICML/vit_schedulers.pdf}
%     \caption{\textbf{ViT-small uncertainty training dynamics on \texttt{CIFAR-10} dataset:} In \textbf{(a)} we use \gls{MCD} (fixed $p=0.5$ both in the embeddings and in the transformer blocks) and we train the model for $200$ epochs with Adam optimizer \citep{Kingma2017} and cosine annealing. In \textbf{(b)} we train the same model for $1000$ epochs and fixed learning rate $10^{-4}$. The training dynamics show that the the shape/speed of convergence of of \gls{TU}, \gls{AU}, \gls{EU} strongly depends on the optimization trajectories underneath.}
%     \label{fig:cifar10-ViT-two-experiments-learning}  
% \end{figure} 

We explore uncertainty scaling trends using Vision Transformer (ViT) architectures \citep{Dosovitskiy2021}. 
We conduct a study to highlight how different experimental settings lead to different uncertainty behaviors (\cref{fig:cifar10-ViT-two-experiments-learning}). It is interesting to observe the effect of annealing the learning rate with a cosine scheduler when training on \texttt{CIFAR-10}, suggesting that early-phase uncertainty dynamics in transformers can be sensitive to optimization strategies. Results on \texttt{ImageNet32} are reported in \cref{app:fig:vit-imagenet32}.

\begin{figure}[t]
    \centering
    \includegraphics[width=0.45\textwidth]{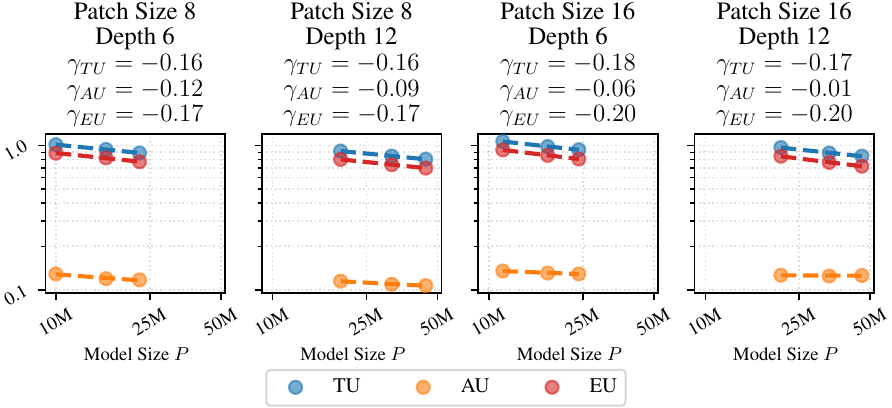}
    \caption{\textbf{ViT uncertainty scaling with model parameters:} We train different sizes of ViT by varying the patch size ($8$, $16$), the number of layers ($6$, $12$) and the number of attention heads (each point corresponds to $8$, $16$ and $24$ attention heads respectively). We use \gls{MCD} (with fixed $p = 0.5$ both in the embeddings and in the transformer blocks). \gls{AU} shows minimal dependence on $P$. \gls{TU} mirrors the trend of the epistemic component. Overall, the slopes are small in magnitude ($|\gamma| < 0.2$), indicating that uncertainty decreases only slowly with model size.}
    \label{fig:vit_scaling}  
\end{figure}

\paragraph{ViT uncertainty scaling with model parameters}
In ViT models, \gls{EU} decreases with model size (negative $\gamma$ across all patch sizes and depths), as it can be observed in \cref{fig:vit_scaling}. We believe that larger ViTs produce smoother representations and show stronger inductive bias toward stable features; consequently, different dropout samples yield increasingly similar predictive distributions. In this setting, model size reduces ensemble diversity, and therefore \gls{EU} contracts.

\subsection{Text classification}
\label{sec:text-domain}

We firstly investigate the language modality by assessing uncertainty estimates on the pre-trained Phi-2 model \citep{Abdin2024}, applying a Laplace approximation to the posterior over the LoRA parameters \citep{Yang2024}. We fine-tuned the model on \texttt{qqp} and \texttt{ARC} datasets observing that the uncertainties remain flat for every data subset used for fine tuning (see \cref{app:fig:bayesian_lora}). This saturation effect is likely due to the massive amount of data used for pre-training, limiting the expressiveness of the model uncertainty on comparatively much smaller data for fine-tuning.

\subsubsection{GPT-2 on Algorithmic Dataset}

\begin{figure}[t]
    \centering \includegraphics[width=0.45\textwidth]{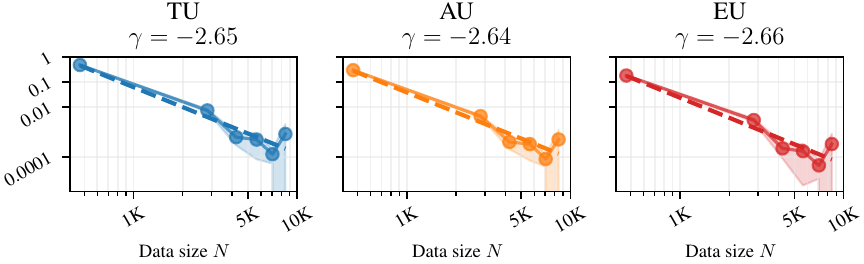}
    \caption{\textbf{Algorithmic dataset}: GPT-2 model on the Algorithmic dataset to learn the modulo addition operation. Results are obtained by training for $10\,000$ epochs and averaging over $3$ seeds.
        %for each considered training subset size: [$5\%$, $30\%$, $45\%$, $60\%$, $75\%$, $90\%$]. 
        %We train this model for $10.000$ epochs with AdamW optimizer and linear scheduler with warmup.
        }
    \label{fig:grok}   
\end{figure} 

% We train a GPT-2 model to solve modular arithmetic problems from a synthetic dataset, following \citet{Power2022}. The model learns to predict the token after the ($=$) sign. We only experiment with \gls{MCD} with rate $p=0.1$. Interestingly, more evident uncertainty scaling patterns only emerge after extensive training, suggesting potential links between grokking dynamics \citep{Belkin2019} and uncertainty behavior. In \cref{fig:grok} we show predictive uncertainties when training the model on increasing percentages (from $5\%$ to $90\%$) of the \texttt{MODULO 97} algorithmic dataset. The slight uncertainty increase in the last $N$ subset of \cref{fig:grok} likely stems from convergence dynamics and limitations of \gls{MCD}. 

We train a GPT-2 model to solve modular arithmetic problems of the form $(a + b) \bmod m =$, where $m$ is the modulus, $a$ and $b$ are integers from $0$ to $m$, and the model learns to predict the token after the $=$ sign. This setup and dataset differ significantly from the vision tasks considered so far, thus providing a complementary testbed for our hypotheses. We experiment using \gls{MCD} with a dropout rate of $p = 0.1$. In \cref{fig:grok} we show predictive uncertainties when training the model on increasing percentages (from $5\%$ to $90\%$) of the $m = 97$ algorithmic dataset. Similarly to previous experiments, we observe clear scaling laws for all uncertainty metrics considered. Interestingly, more evident uncertainty scaling patterns only emerge after extensive training, suggesting potential links between grokking \citep{Belkin2019} and uncertainty behavior. The slight uncertainty increase in the last $N$ subset of \cref{fig:grok} likely stems from convergence dynamics and limitations of \gls{MCD}.

\subsection{Extrapolation on a toy dataset}
\label{sec:mlp-extrapolation}

\begin{figure}[t]
    \centering
    \includegraphics[width=0.45\textwidth]{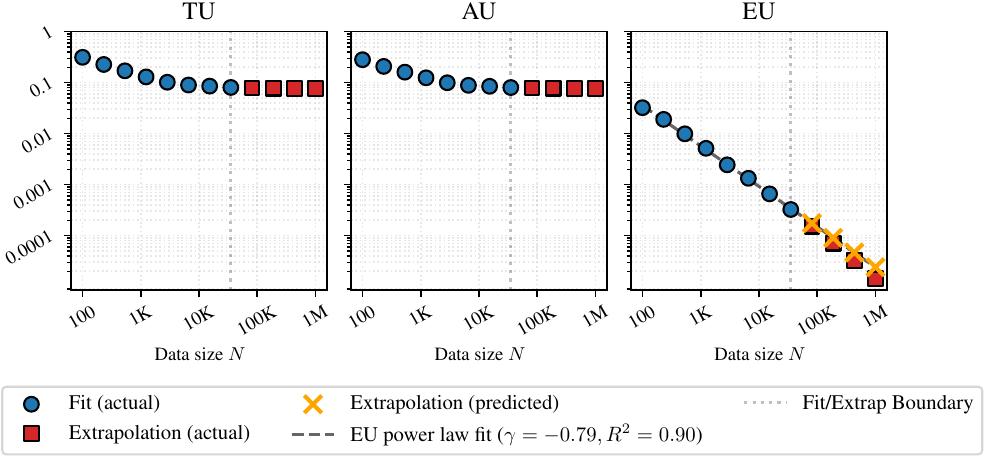}
    \caption{\textbf{Uncertainty scaling of an \gls{MLP} on the \texttt{two\_moons} dataset}: We use Bayesian logistic regression with normal priors on an \gls{MLP} with two hidden layers of size $32$ and $\tanh$ activation functions. Posterior inference is performed with \gls{HMC} sampling ($200$ posterior samples after $1{,}000$ warm-up steps). Training is carried out on $12$ log-spaced values ranging from $100$ to $1M$ and $4$ random seeds. Uncertainties are computed on the held-out $5{,}000$ test set.}
    \label{fig:mlp-etrapolation}  
\end{figure}

We conduct an experiment using a simple \gls{MLP} on the \texttt{two\_moons} \footnote{Dataset available through Scikit-learn \citep{scikit-learn}} toy classification dataset to demonstrate the practical application of uncertainty scaling laws. As illustrated in \cref{fig:mlp-etrapolation}, the \gls{EU} exhibits a clear power-law decay across the observed range of $N$. As expected, the \gls{AU} contribution settles down and plateaus for large values of $N$, representing the irreducible noise inherent in the task. The \gls{TU} curve eventually diverges from the power-law trend as it approaches this aleatoric floor. Beyond the extrapolation boundary, the predicted \gls{EU} remains highly consistent with the actual observed scaling. This is particularly powerful because it demonstrates that training on a relatively small regime, up to just a few thousand data points, allows us to accurately predict the uncertainty behavior we can expect when scaling the data to the order of millions of samples.

\section{Theoretical connections}

The theory of identifiable models has extensively analyzed asymptotic behaviors, such as posterior contraction and convergence of test loss and generalization error. In this section, we recall how predictive uncertainty contracts with increasing data in Bayesian linear regression, highlight its connection to Watanabe’s Generalization Error, and outline links to \gls{SLT}.

\subsection{Total Uncertainty scaling in Identifiable Parametric Models}
\label{sec:theory:tu_scaling}

We consider Bayesian linear regression $y = \mbtheta^\top\mbphi(\mbx) + \epsilon$, where $\mbtheta \in \bbR^P$ are parameters of interest, and $\mbphi(\mbx) = [\phi_1(\mbx),\phi_2(\mbx),\dotsc,\phi_P(\mbx)]^\top$ are basis functions, and assume zero-mean noise $\epsilon \sim \N(0, \sigma^2)$. We define the likelihood over $N$ iid observations $\{\mbX, \mby\} = \{(\mbx_i, y_i)\}_{i=1}^N$,
\begin{align}
    p(\mby|\mbX,\mbtheta,\sigma^2) &= \prod_{n=1}^N \cN\big(y_n|\mbphi(\mbx_n)^\top\mbtheta, \sigma^2\big).
\end{align}
By assuming a conjugate Gaussian prior $p(\mbtheta) = \cN(\mbtheta|\mbm_0,\mbS_0)$ the posterior is also Gaussian with mean $\mbm_N = \mbS_N\left(\mbS_0^{-1}\mbm_0 + \sigma^{-2} \mbPhi^\top\mby \right)$ and covariance $\mbS_N = \left( \sigma^{-2}\mbPhi^\top\mbPhi + \mbS_0^{-1}\right)^{-1}$.

The predictive posterior for a new test points $(\mbx_*,y_*)$ is also Gaussian,
$    p(y_*|\mbx_*,\mby,\mbX) = \cN\Big(y_*|\mbm_N^\top\mbphi(\mbx_*), \sigma^2_N(\mbx_*)\Big) $
with $\sigma^2_N(\mbx_*) = \sigma^2 + \mbphi(x_*)^\top\mbS_N\mbphi(x_*). 
    \label{eq:blr-predictive}
$

The predictive variance $\mathrm{Var}\Big[y_*\mid\mbx_*,\mby,\mbX\Big]$ decomposes into the uncertainty of data  $\sigma^2$ (\gls{AU}), and the uncertainty of parameters $\mbphi(\mbx_*)^\top\mbS_N\mbphi(\mbx_*)$ (\gls{EU}). 
%We stress that \gls{EU} is made available only by taking the Bayesian approach of treating $\mbtheta$ as a random variable.
It can be shown that $\sigma^2_{N+1}(\mbx_*) \le \sigma^2_N(\mbx_*)$ -- the posterior distribution becomes narrower as additional data points are observed \citep{Qazaz1997}. Moreover, in the limit of $N \rightarrow \infty$ we get $\sigma^2_N(\mbx) \rightarrow \sigma^2$: the predictive uncertainty converges to its irreducible \gls{AU} component.

\subsection{Singular Learning Theory}

Our speculative theoretical link is \gls{SLT}, which provides insights into the learning dynamics of highly overparameterized models. The Fisher information of deep neural networks is often singular at certain parameter values; despite forming a measure-zero subset, these singularities significantly influence learning. \gls{SLT} proves that the asymptotic properties of learning are shaped by the geometry near such degenerate points \citep{Watanabe2009, Watanabe2018}. 
%One main quantity of interest in this framework is the generalization error $G$. 
Here we limit ourself to connect this to the \gls{TU} in Bayesian linear regression. Using Watanabe's notation: 
\begin{equation}
    F_N = -\log p(\mby_N|\mbX_N) = - \log \int p(\mby_N | \mbX_N, \mbtheta) p(\mbtheta) d\mbtheta 
\end{equation}
is the \gls{NMLL} of the data $\mbX_N$ of size $N$. By denoting the true conditional distribution $p(y) \equiv p(y_{N+1} | \mbx_{N+1}, \mbtheta_\mathrm{true})$ (e.g., $y_{N+1} \sim \cN\big(\mbtheta_\mathrm{true}^\top\mbx_{N+1}, \sigma_\mathrm{true}^2\big)$), the generalization error $G_N = \E_{p(y)}[F_{N+1}] - F_N$ measures the expected increase in marginal likelihood when training with one additional data point (see \cref{sec:app:watanabe}). This can be rewritten as:
\begin{equation} 
    % G_N &= \E_{p(y_{N+1}|\mbx_{N+1},\hat{\mbtheta})}[F_{N+1}] - F_N
     G_N =  -\E_{p(y)} \Big[ \log \underbrace{p\big(y_{N+1} | \mbx_{N+1}, \mbX_N, \mby_N\big)}_{q_N(y)} \Big], \label{eq:watanabe-generalizazation-main}
         %&= \E_{p(y)}[\log q(y)]
\end{equation}

which reduces to the log posterior predictive distribution for the $(N+1)$'th datapoint given the first $N$ datapoints, under the true conditional distribution $p(y)$. By denoting the posterior predictive $q_N(y)$, we can manipulate the expression in \cref{eq:watanabe-generalizazation-main} to obtain:

\begin{equation}
    G_N = \underbrace{\frac{1}{2} \log{(2\pi e \sigma_\mathrm{true}^2)}}_{\text{aleatoric uncertainty}} + \underbrace{\mathrm{KL}\big[p(y) || q_N(y)\big]}_{\text{epistemic uncertainty}},
\end{equation}
where $\mathrm{KL}[p(y) || q_N(y)]$ is the \gls{KL} divergence between the true and the posterior predictive distribution, quantifying the \gls{EU} arising from limited knowledge of the model parameters. 
By looking at the asymptotic expression of \gls{TU} as $N \rightarrow \infty$ and by taking, without loss of generality, identity basis functions, we get:
\begin{equation}
\begin{split}
    \mathrm{TU}(\mbx_{N+1}) &= \mathbb{H}[q_N(y)] \\
    &= \frac{1}{2}\log(2\pi e \sigma_\mathrm{true}^2) + \frac{\mbx_{N+1}^\top\mbSigma_{X_{N+1}}^{-1}\mbx_{N+1}}{2(N+1)} \\
    &\quad + O\left(\frac{1}{(N+1)^2}\right). 
\end{split}
\end{equation}

As more data is collected, the predictive posterior $q_N(y)$ converges to the true predictive distribution $p(y)$. Consequently, the generalization error $G_N$ asymptotically approaches the irreducible \gls{AU}. Similarly, the \gls{TU} converges to the aleatoric component as the epistemic part vanishes. In \citet{Watanabe1999} it's proved an asymptotic expansion of $G_N$ which is proportional to the effective dimensionality of the model (\cref{sec:watanabe_asymptotic_expansion}).

Recent advances on the effective dimensionality of deep models \citep{Lau2024, Chen2024} appear especially promising, and we intend to investigate such formal connections in future work. We further hypothesize that concepts from Statistical Physics may help explain the scaling behaviors we observed.

\section{Conclusions}

Inspired by recent work on scaling laws in deep learning, we investigated whether similar patterns hold for predictive uncertainties. Across vision and language tasks, we find scaling behaviors robust to architecture, posterior approximation, and hyper-parameters. Building on \gls{SLT}, we provide theoretical insights connecting generalization error and predictive uncertainty, showing why information-theoretic measures scale with dataset size. However, deriving exact power-law coefficients is difficult, as they vary unpredictably with design choices.
Our results suggest practical strategies for extrapolating uncertainties to large $N$, such as estimating how much data is needed for ensemble predictions to converge. They also point to applications in active learning, where one can predict the marginal uncertainty reduction from additional data to guide annotation budgets. More broadly, while scaling laws are predictive, they are not universal but depend on how the loss landscape is traversed, underscoring the need to tailor optimization strategies to each task. 

% Future work will explore alternative theories, such as linking uncertainties in wide networks with \glspl{GP} limits. We also plan to examine scaling with compute budget and model size more in depth.

\paragraph{Limitations and future work}
A key limitation is that we do not yet have a formal theory that predicts the empirical scaling exponents $\gamma$ of deep networks from \gls{SLT} invariants or other geometric quantities. Moreover, optimization effects remain only partially characterized: we observe that simple optimization settings (e.g., fixed learning rates) can yield clearer scaling than more complex schedules in some settings, but we do not yet explain when and why this occurs. Finally, in transfer and fine-tuning settings, large-scale pre-training can saturate uncertainty estimates, potentially masking scaling behavior on smaller downstream datasets.

Future work will aim to (i) connect deep-network scaling exponents to effective model complexity (e.g., effective dimensionality or RLCT-like quantities), (ii) study more in depth the scaling with compute budget and model size, and (iii) relate uncertainty scaling to theoretical limits such as wide-network \gls{GP} approximations.

\section*{Impact Statement}

By characterizing uncertainty scaling in modern neural networks, this work provides essential insights for ensuring the reliability and safety of large-scale AI systems.


\bibliography{bibliography}
\bibliographystyle{icml2026}

%%%%%%%%%%%%%%%%%%%%%%%%%%%%%%%%%%%%%%%%%%%%%%%%%%%%%%%%%%%%%%%%%%%%%%%%%%%%%%%
%%%%%%%%%%%%%%%%%%%%%%%%%%%%%%%%%%%%%%%%%%%%%%%%%%%%%%%%%%%%%%%%%%%%%%%%%%%%%%%
% APPENDIX
%%%%%%%%%%%%%%%%%%%%%%%%%%%%%%%%%%%%%%%%%%%%%%%%%%%%%%%%%%%%%%%%%%%%%%%%%%%%%%%
%%%%%%%%%%%%%%%%%%%%%%%%%%%%%%%%%%%%%%%%%%%%%%%%%%%%%%%%%%%%%%%%%%%%%%%%%%%%%%%
\newpage
\appendix
\onecolumn

\include{appendix}
%%%%%%%%%%%%%%%%%%%%%%%%%%%%%%%%%%%%%%%%%%%%%%%%%%%%%%%%%%%%%%%%%%%%%%%%%%%%%%%
%%%%%%%%%%%%%%%%%%%%%%%%%%%%%%%%%%%%%%%%%%%%%%%%%%%%%%%%%%%%%%%%%%%%%%%%%%%%%%%

\end{document}

%% file: config/packages.tex
\usepackage[utf8]{inputenc} 
\usepackage[T1]{fontenc}    
\usepackage{hyperref}      
\usepackage{url}            
\usepackage{booktabs}       
\usepackage{amsfonts}      
\usepackage{verbatim}       
\usepackage{nicefrac}      
\usepackage{microtype}      
\usepackage{xcolor}         
\usepackage{comment}
\usepackage{wrapfig}
\usepackage{amssymb}
\usepackage{mathtools}
\usepackage{amsmath}
\usepackage{natbib}
\usepackage{amsthm}
\usepackage[capitalize,noabbrev]{cleveref}
\usepackage{graphicx}
\usepackage{subcaption}

\usepackage{tcolorbox}

\definecolor{mylightgray}{gray}{0.95}
\definecolor{boxcolor}{HTML}{faf9f5}
\newtcolorbox{mybox}{colback=mylightgray,colframe=mylightgray,top=0.8pt,bottom=0.8pt,right=1.8pt,left=1.8pt}

\usepackage[textsize=tiny]{todonotes}

\hypersetup{colorlinks,citecolor=blue!50!green,linkcolor=red!80!black,urlcolor=blue}
\usepackage[all]{hypcap}
\usepackage{xspace}
\usepackage[acronym,nowarn]{glossaries}
\glsdisablehyper
\makeglossaries
\usepackage{pdfpages}
\usepackage{float}

\usepackage{tikz,siunitx}
\usetikzlibrary{shapes.geometric,shapes.symbols}

%% file: config/math.tex
\newcommand{\mathbold}[1]{{\boldsymbol{\mathbf{#1}}}}

\newcommand{\nestedmathbold}[1]{{\mathbold{#1}}}

\def\0{\mathbf{0}}

\def\R{\mathbb{R}}
\def\N{\mathcal{N}}

\newcommand{\mbf}{\nestedmathbold{f}}

\newcommand{\mbm}{\nestedmathbold{m}}

\newcommand{\mbp}{\nestedmathbold{p}}

\newcommand{\mbx}{\nestedmathbold{x}}
\newcommand{\mby}{\nestedmathbold{y}}

\newcommand{\mbI}{\nestedmathbold{I}}

\newcommand{\mbS}{\nestedmathbold{S}}

\newcommand{\mbX}{\nestedmathbold{X}}

\newcommand{\mbmu}{\nestedmathbold{\mu}}

\newcommand{\mbphi}{\nestedmathbold{\phi}}

\newcommand{\mbtheta}{\nestedmathbold{\theta}}

\newcommand{\mbLambda}{\nestedmathbold{\Lambda}}

\newcommand{\mbPhi}{\nestedmathbold{\Phi}}

\newcommand{\mbSigma}{\nestedmathbold{\Sigma}}

\newcommand{\mbzero}{\nestedmathbold{0}}

\DeclarePairedDelimiterX{\infdivx}[2]{[}{]}{%
  #1\;\delimsize\|\;#2%
}

\newcommand{\cD}{\mathcal{D}}

\newcommand{\cN}{\mathcal{N}}

\newcommand{\E}{\mathbb{E}}
\newcommand{\V}{\mathbb{V}}
\newcommand{\bbH}{\mathbb{H}}
\newcommand{\bbR}{\mathbb{R}}

%% file: config/acronyms.tex
\newacronym{PSD}{psd}{Positive Semi-Definite}

\newacronym{OOD}{\textsc{ood}}{out-of-distribution}

\newacronym{MNLL}{mnll}{mean negative loglikelihood}
\newacronym{NLL}{nll}{negative loglikelihood}
\newacronym{LL}{ll}{log-likelihood}
\newacronym{RMSE}{rmse}{root mean square error}
\newacronym{ECE}{ece}{expected calibration error}
\newacronym{GGN}{\textsc{ggn}}{Generalized Gauss-Newton}
\newacronym{MAP}{\textsc{map}}{Maximum a posteriori}
\newacronym{GLM}{\textsc{glm}}{Generalized linear model}

\newacronym{MC}{\textsc{mc}}{Monte Carlo}
\newacronym{MCD}{\textsc{mcd}}{Monte Carlo Dropout}
\newacronym{DE}{\textsc{de}}{Deep Ensembles}
\newacronym{MCMC}{\textsc{mcmc}}{Markov chain Monte Carlo}
\newacronym{HMC}{\textsc{hmc}}{Hamiltonian Monte Carlo}
\newacronym{SGHMC}{\textsc{sghmc}}{Stochastic Gradient Hamiltonian Monte Carlo}
\newacronym{SGLD}{\textsc{sgld}}{Stochastic Gradient Langevin Dynamics}
\newacronym{LLA}{\textsc{lla}}{Linearized Laplace Approximation}
\newacronym{GP}{\textsc{gp}}{Gaussian process}

\newacronym{VI}{\textsc{vi}}{Variational Inference}

\newacronym{ELBO}{\textsc{elbo}}{evidence lower bound}
\newacronym{ELL}{\textsc{ell}}{expected log likelihood}
\newacronym{KL}{\textsc{kl}}{Kullback-Leibler divergence}
\newacronym{AUC}{\textsc{auc}}{area under the curve}
\newacronym{MI}{\textsc{mi}}{Mutual Information}

\newacronym[firstplural=Bayesian neural networks]{BNN}{\textsc{bnn}}{Bayesian neural network}
\newacronym[firstplural=deep neural networks]{DNN}{\textsc{dnn}}{deep neural network}
\newacronym[]{CNN}{\textsc{cnn}}{convolutional neural network}
\newacronym{MLP}{\textsc{mlp}}{multilayer perceptron}
\newacronym{NN}{\textsc{nn}}{neural network}
\newacronym{RELU}{\textsc{ReLU}}{rectified linear unit}

\newacronym{WD}{wd}{Wasserstein distance}

\newacronym{NTK}{ntk}{Neural Tangent Kernel}

\newacronym{ESS}{ess}{Effective Sample Size}

\newacronym{AU}{\textsc{au}}{Aleatoric Uncertainty}
\newacronym{EU}{\textsc{eu}}{Epistemic Uncertainty}
\newacronym{TU}{\textsc{tu}}{Total Uncertainty}
\newacronym{UQ}{\textsc{uq}}{Uncertainty Quantification}
\newacronym{SLT}{\textsc{slt}}{Singular Learning Theory}
\newacronym{SAM}{\textsc{sam}}{Sharpness Aware Minimization}
\newacronym{SGD}{\textsc{sgd}}{Stochastic Gradient Descent}
\newacronym{IVON}{\textsc{ivon}}{Improved Variational Online Newton}
\newacronym{RLCT}{\textsc{rlct}}{Real Log Canonical Threshold}
\newacronym{NMLL}{\textsc{nmll}}{Negative Marginal Log Likelihood}
\newacronym{LLC}{\textsc{llc}}{Local Learning Coefficient}
\newacronym{LLM}{\textsc{llm}}{Large Language Model}

%% file: config/theorems.tex
\theoremstyle{plain}

\theoremstyle{definition}

\theoremstyle{remark}

%% file: appendix.tex
\section{Connections with \gls{SLT}}

Our theoretical framework is motivated by \gls{SLT}, which offers geometric insights into the learning dynamics of highly overparameterized models. In deep neural networks, the Fisher information matrix is frequently singular; although these singularities form a measure-zero subset of the parameter space, they profoundly influence the learning process. \gls{SLT} demonstrates that the asymptotic properties of learning are governed by the algebraic geometry near these degenerate points \citep{Watanabe2009, Watanabe2018}.

In this section, we restrict our analysis to the connection between \gls{SLT} and \gls{TU} within the context of Bayesian linear regression. Following the notation of \citet{Watanabe2009}, we define the \gls{NMLL} (or free energy) of a dataset $\mbX_N$ of size $N$ as:
\begin{equation}
    F_N = -\log p(\mby_N|\mbX_N) = - \log \int p(\mby_N | \mbX_N, \mbtheta) p(\mbtheta) d\mbtheta.
\end{equation}
The generalization error, $G_N$, measures the expected negative log-likelihood of an unseen data point $(\mbx_{N+1}, y_{N+1})$ sampled from the true distribution. It is defined as:
\begin{align} 
    G_N &= \E_{p(y_{N+1}|\mbx_{N+1},\mbtheta_{\text{true}})} \left[ -\log p(y_{N+1} | \mbx_{N+1}, \mbX_N, \mby_N) \right] \label{eq:watanabe-generalizazation-main-appendix} \\
        &= -\E_{p(y)} \left[ \log q_N(y) \right],
\end{align}
where $p(y)$ denotes the true process $y_{N+1} \sim \cN(y_{N+1}|\mbtheta_\mathrm{true}^\top\mbx_{N+1}, \sigma_\mathrm{true}^2)$ and $q_N(y)$ represents the posterior predictive distribution given the first $N$ data points.

We can decompose the generalization error in \cref{eq:watanabe-generalizazation-main-appendix} into aleatoric and epistemic components:
\begin{align}
    G_N &= \E_{p(y)}[-\log{q_N(y)}] \nonumber \\
        &= \E_{p(y)}[-\log p(y) + \log p(y) - \log{q_N(y)}] \nonumber \\
        &= \underbrace{\mathbb{H}[p(y)]}_{\text{aleatoric uncertainty}} + \underbrace{\mathrm{KL}\big[p(y) || q_N(y)\big]}_{\text{epistemic uncertainty}}. \label{eq:GN_decomposition}
\end{align}
Here, $\mathbb{H}[p(y)] = \frac{1}{2} \log{(2\pi e \sigma_\mathrm{true}^2)}$ represents the irreducible noise (entropy) of the true process, while the \gls{KL} divergence quantifies the \gls{EU} arising from the deviation of the posterior predictive from the truth.

Examining the asymptotic behavior of \gls{TU} as $N \rightarrow \infty$, and assuming identity basis functions without loss of generality, we obtain:
\begin{equation}
    \mathrm{TU}(\mbx_{N+1}) = \mathbb{H}[q_N(y)] 
    = \frac{1}{2}\log(2\pi e \sigma_\mathrm{true}^2) + \frac{\mbx_{N+1}^\top\mbSigma_{X_{N+1}}^{-1}\mbx_{N+1}}{2(N+1)} + O\left(\frac{1}{N^2}\right). 
\end{equation}
As $N \to \infty$, the posterior predictive $q_N(y)$ converges to the true distribution $p(y)$. Consequently, the epistemic term vanishes, and both $G_N$ and \gls{TU} asymptotically approach the irreducible aleatoric entropy. \citet{Watanabe1999} derives a rigorous asymptotic expansion of $G_N$, showing it is proportional to the effective dimensionality of the model (see \cref{sec:watanabe_asymptotic_expansion}).

Recent advances in estimating the effective dimensionality of deep models \citep{Lau2024, Chen2024} appear especially promising. We intend to investigate these formal connections in future work, hypothesizing that concepts from Statistical Physics may further elucidate the observed scaling behaviors.

\subsection{Total Uncertainty in Bayesian Linear Regression}
\label{sec:app:blr}

%We begin with a simplified setting which has well-known properties that are useful to define some quantities of interest for this problem. 

We consider a class of models defined as a linear combination of fixed nonlinear functions of the input variables of the form:

\begin{equation}
        y(\mbx,\mbtheta) = \sum_{j=0}^{M-1}\theta_j\phi_j(\mbx) = \mbtheta^\top\mbphi(\mbx),
\end{equation}

where $\mbtheta = (\theta_0,\dotsc,\theta_{M-1})^\top$ are model parameters and $\mbphi = (\phi_0,\dotsc,\phi_{M-1})^\top$ are known as \textit{basis functions} which allow $y(\mbx,\mbtheta)$ to be a non-linear function of the input $\mbx$. 

We consider the target variable $y$ that is given by a deterministic function $y(\mbx,\mbtheta)$ plus additive Gaussian noise:

\begin{equation}
y = \mbtheta^\top\mbphi(\mbx) + \epsilon \text{,}
\end{equation}

where $\mbtheta$ is treated as a random vector and $\epsilon \sim \cN(0,\sigma^2)$, obtaining

\begin{equation}
        p(y|\mbx,\mbtheta,\sigma^2) = \cN(y|\mbphi(\mbx)^\top\mbtheta, \sigma^2) \text{.}
    \label{blr:eq:blr-likelihood}
\end{equation}

We consider a supervised learning problem with $N$ input-label training pairs $\{\mbX,\mby\} = \{(\mbx_i,y_i)\}_{i=1}^N$. Assuming that the data points are drawn independently from \cref{blr:eq:blr-likelihood}, the likelihood becomes:

\begin{equation}
        p(\mby|\mbX,\mbtheta,\sigma^2) = \prod_{n=1}^N \cN(y_n|\mbphi(\mbx_n)^\top\mbtheta, \sigma^2) \text{.}
\end{equation}

According to the transformation of the features introduced by the set of basis functions $\mbphi$, we define the design matrix $\mbPhi \in \bbR^{N \times M}$ with entries $\mbPhi_{nj} = \phi_j(\mbx_n)$. We assume a conjugate prior over $\mbtheta$:

\begin{equation}
        p(\mbtheta) = \cN(\mbtheta|\mbmu_0,\mbSigma_0) \text{.}
\end{equation}

From Bayes' theorem, the posterior can be derived in closed form leading to the following result:

    \begin{align}
        p(\mbtheta|\mby,\mbX) &= \cN(\mbtheta|\mbmu_N, \mbSigma_N) \text{,}\\
        \mbmu_N &=\mbSigma_N\left(\mbSigma_0^{-1}\mbmu_0 + \sigma^{-2}\mbPhi^\top\mby\right) \text{,} \\
        \mbSigma_N &= \left( \sigma^{-2}\mbPhi^\top\mbPhi + \mbSigma_0^{-1}\right)^{-1} \text{.}
    \end{align}

We are interested in the predictive uncertainty for a new test point $(\mbx_*,y_*)$ so we inspect the predictive posterior and its properties:

    \begin{align}
        p(y_*|\mbx_*,\mby,\mbX) &= \int p(y_*|\mbx_*,\mbtheta)p(\mbtheta|\mby,\mbX)\text{d}\mbtheta  \nonumber \\
        &= \cN(y_*|\mbmu_N^\top\mbphi(\mbx_*), \underbrace{\sigma^2 + \mbphi(\mbx_*)^\top\mbSigma_N\mbphi(\mbx_*)}_{\V[y_*|\mbx_*,\mby]}) \text{,}
        \label{eq:blr-predictive}
    \end{align}

where $\mbmu_N$ and $\mbSigma_N$ are the posterior mean and covariance matrix. From the predictive variance $\V[y_*|\mbx_*,\mby]$ we identify two components: (i) $\sigma^2$ which represents the uncertainty related to the data noise (\gls{AU}) and (ii) $\mbphi(\mbx_*)^\top\mbSigma_N\mbphi(\mbx_*)$ which is the uncertainty associated with model parameters (\gls{EU}). As we discuss in \cref{sec:theory:tu_scaling}, by defining $\V[y_*|\mbx_*,\mby] \coloneq \sigma^2_N(\mbx)$ as the predictive uncertainty for a new test point $\mbx$ when training on $N$ input points, we have that $\sigma^2_{N+1}(\mbx) \le \sigma^2_N(\mbx)$, meaning that the posterior distribution becomes narrower as additional data points are observed \citep{Qazaz1997}. Moreover, $\lim_{N\to\infty} \sigma^2_N(\mbx) = \sigma^2$; that is, in the infinite data regime, the predictive uncertainty converges to its irreducible component, \gls{AU}.

\subsubsection{From posterior predictive variance to Total Uncertainty}
\label{sec:app:tu}

We now derive an expression for the (total) predictive uncertainty associated with a new data point $(\mbx,y)$, defined as the entropy of the predictive posterior:

    \begin{align}
        \text{TU}(\mbx) &= \bbH\left[p(y|\mbx,\mby,\mbX)\right] \nonumber \\
        &= \frac{1}{2}\log\left(2\pi e \sigma_N^2(\mbx)\right) \nonumber \\
        &= \frac{1}{2}\log\left(2\pi e \left(\sigma^2 + \mbphi(\mbx)^\top\left( \sigma^{-2}\mbPhi^\top\mbPhi + \mbSigma_0^{-1}\right)^{-1}\mbphi(\mbx)\right)\right) \nonumber \\
        &=\frac{1}{2}\log\left(2\pi e \sigma^2\left(1 + \sigma^{-2}\mbphi(\mbx)^\top\left(\sigma^{-2}\mbPhi^\top\mbPhi + \mbSigma_0^{-1}\right)^{-1}\mbphi(\mbx)\right)\right) \nonumber \\
        &= \frac{1}{2}\log\left(2\pi e \sigma^2\right) + \frac{1}{2}\log\left(1 + \sigma^{-2}\mbphi(\mbx)^\top\left( \sigma^{-2}\mbPhi^\top\mbPhi + \mbSigma_0^{-1}\right)^{-1}\mbphi(\mbx)\right) \nonumber \\
        &= \underbrace{\frac{1}{2}\log\left(2\pi e \sigma^2\right)}_{\text{AU}(\mbx)} + \underbrace{\frac{1}{2N}\mbphi(\mbx)^\top\mbSigma_{\phi}^{-1}\mbphi(\mbx)}_{\text{EU}(\mbx)} + \, O\left(\frac{1}{N^2}\right)
    \end{align}

In the previous derivation, we assume $\mbX = (\mbx_1,\dotsc,\mbx_N)$ to be a collection of IID samples with $\mbzero$ mean and $\mbSigma_x$ covariance matrix. The same holds for $\mbPhi$ where samples $\mbphi(\mbx)$ are drawn independently from a distribution with $\mbzero$ mean and $\mbSigma_\Phi$ covariance matrix. By the Law of Large Numbers $\frac{1}{N}\mbPhi^\top\mbPhi \rightarrow \E[\mbphi(\mbx)\mbphi(\mbx)^\top] = \mbSigma_\Phi$ as $N \rightarrow \infty$ which implies $\mbPhi^\top\mbPhi \rightarrow N\mbSigma_\Phi$. In addition, for large $N$ the $\mbSigma_0^{-1}$ term vanishes. This shows that as $N \rightarrow \infty$, \gls{TU} approaches \gls{AU} and the \gls{EU} decays with rate $\frac{1}{N}$.

\subsection{Total Uncertainty and Generalization Error for linear models}
\label{sec:app:watanabe}

In this Section we formally derive a connection between the Watanabe generalization error $G_N$ and the Total Uncertainty (\gls{TU}) for Bayesian linear models. Using the notation from \citet{Watanabe2009,Watanabe2018} the generalization error $G_N$ and the Free Energy $F_N$ are defined as:

\begin{align}
    G_N &= \E_{p(y_{N+1}|\mbx_{N+1},\mbtheta_{\mathrm{true}})}[F_{N+1}] - F_N \\
    F_N &= -\log p(\mby_N|\mbX_N) = - \log \int p(\mby_N | \mbX_N, \boldsymbol\theta) p(\boldsymbol\theta) d\boldsymbol\theta .
\end{align}

We consider the same linear model of \cref{sec:app:blr} and without loss of generality we set the basis functions to be identity functions, i.e., $\mbphi(\mbx) = \mbx$. We also introduce $\mbtheta_{\mathrm{true}}$, the set of underlying true parameters. A new data point $(\mbx_{N+1},y_{N+1})$ is generated according to the true process:
\begin{equation}
    y_{N+1} \sim \cN(y_{N+1}|\mbx_{N+1},\mbtheta_{\mathrm{true}}) = \cN(y_{N+1}|\mbtheta_{\mathrm{true}}^\top\mbx_{N+1}, \sigma_{\mathrm{true}}^2).
\end{equation}

The free energy $F_N$ is defined as the negative log marginal likelihood:

\begin{align}
    F_N &= - \log{p(\mby_N|\mbX_N)} \nonumber \\
    &= - \log{\int{p(\mby_N|\mbX_N,\mbtheta)p(\mbtheta)d\mbtheta}} \nonumber \\
    &= - \log{\int{\left(\prod_{n=1}^Np(y_n|\mbx_n,\mbtheta)\right)p(\mbtheta)d\mbtheta}}.
\end{align}

% The posterior distribution is in closed form:

% \begin{align}
%     p(\mbtheta|\mbmu_N,\mbSigma_N)
% \end{align}

% where $\mbSigma_N^{-1} = \Sigma_0^{-1} + \beta\mbX_N^\top \mbX_N$ and $\mbmu_N = \mbSigma_N(\mbSigma_0^{-1}\mbmu_0 + \beta\mbX_N^\top \mby_N)$.

% The generalization error we want to compute is:

% \begin{align}
%     G_N &= \E_{p(y_{N+1}|\mbx_{N+1},\hat{\mbtheta})}[F_{N+1} - F_N]\\
%     &= \E_{p(y_{N+1}|\mbx_{N+1},\hat{\mbtheta})}[F_{N+1}] - F_N
%     \label{eq:watanabe-generalizazation}
% \end{align}

We now relate $F_{N+1}$ and $F_N$, the free energy for $N+1$ and $N$ input points, respectively:

\begin{align}
    F_{N+1} &= -\log p(Y_{N+1}|\mbX_{N+1})\nonumber \\
    &= -\log p(y_{N+1},\mby_N|\mbx_{N+1},\mbX_N) \nonumber \\
    &= -\log p(y_{N+1}|\mbx_{N+1},D_N)p(\mby_N|\mbX_N) \nonumber \\
    &= -\log p(y_{N+1}|\mbx_{N+1},D_N) -\log p(\mby_N|\mbX_N) \nonumber \\
    &= -\log p(y_{N+1}|\mbx_{N+1},D_N) + F_N
\end{align}

where $D_N = \{\mbX_N,\mby_N\}$ and $p(y_{N+1}|\mbx_{N+1},D_N)$ is the posterior predictive distribution for the $(N+1)$ data point given the first $N$ data points:

\begin{align}
    p(y_{N+1}|\mbx_{N+1},D_N) &= \int{p(y_{N+1}|\mbx_{N+1},\mbtheta)p(\mbtheta|D_N)d\mbtheta} \nonumber \\
    &= \cN(y_{N+1}|\mu_{\mathrm{pred}}, \sigma^2_{\mathrm{pred}}),  \\
    \mu_{\mathrm{pred}} &:= \E[y_{N+1}|\mbx_{N+1},D_N] = \mbmu_N^\top\mbx_{N+1} \\
    \sigma^2_{\mathrm{pred}} &:= \V[y_{N+1}|\mbx_{N+1},D_N] = \sigma_{\mathrm{true}}^2 + \mbx_{N+1}^\top \mbSigma_N\mbx_{N+1}.
\end{align}

Let's simplify the notation: we rename the new data point $y_{N+1}$ simply $y$ such that $p(y) \coloneq p(y_{N+1}|\mu_{\mathrm{true}},\sigma_{\mathrm{true}}^2)$ where $\mu_{\mathrm{true}} \coloneq \mbtheta_{\mathrm{true}}^\top \mbx_{N+1}$. We can now compute the generalization error:

% \begin{align}
%     \E_{p(y)}[F_{N+1}] &= \E_{p(y)}[F_N - \log{p(y_{N+1}|\mbx_{N+1},D_N)}]\\
%     &= F_N + \E_{p(y)}[-\log{p(y_{N+1}|\mbx_{N+1},D_N)}] \\
%     &= F_N + \E_{p(y)}\left[-\frac{1}{2}\log{(2\pi\sigma_{\mathrm{pred}}^2)} - \frac{(y-\mu_{\mathrm{pred}})^2}{2\sigma_{\mathrm{pred}}^2}\right]
% \end{align}

% where $\E_{p(y)}[(y-\mu_{\mathrm{pred}})^2] = \V_{p(y)}[y] + (\mu_{\mathrm{true}}-\mu_{\mathrm{pred}})^2 = \sigma^2 + \left(\mbtheta_{\mathrm{true}}^\top \mbx_{N+1} - \mbmu_N^\top\mbx_{N+1}\right)^2$. Plugging in the values for $\mu_{\mathrm{pred}}$ and $\sigma_{\mathrm{pred}}^2$ yields to:

\begin{align}
    G_N &= \E_{p(y)}[F_{N+1}] - F_N \nonumber \\
    &= \E_{p(y)}[F_{N+1} - F_N] \nonumber \\
    &= \E_{p(y)}[-\log{p(y_{N+1}|\mbx_{N+1},D_N)}] \nonumber \\
    &= \E_{p(y)}\left[\frac{1}{2}\log{(2\pi\sigma_{\mathrm{pred}}^2)} + \frac{(y-\mu_{\mathrm{pred}})^2}{2\sigma_{\mathrm{pred}}^2}\right] \nonumber \\
    &= \frac{1}{2}\log{(2\pi\sigma_{\mathrm{pred}}^2)} + \frac{\sigma_{\mathrm{true}}^2 + (\mu_{\mathrm{true}}-\mu_{\mathrm{pred}})^2}{2\sigma_{\mathrm{pred}}^2} \nonumber \\
    &= \underbrace{\bbH[\cN(y|\mu_{\mathrm{true}},\sigma_{\mathrm{true}}^2)]}_{\text{AU}(\mbx_{N+1})} + \underbrace{\text{KL}\left[\cN(y|\mu_{\mathrm{true}},\sigma_{\mathrm{true}}^2)||\cN(y|\mu_{\mathrm{pred}},\sigma_{\mathrm{pred}}^2)\right]}_{\text{EU}(\mbx_{N+1})}
    %&= \frac{1}{2}\log{(2\pi\sigma_{\mathrm{pred}}^2)} + \frac{\sigma^2 + \left(\mbtheta_{\mathrm{true}}^\top \mbx_{N+1} - \mbmu_N^\top \mbx_{N+1}\right)^2}{2\sigma_{\mathrm{pred}}^2} \\
    %&= \frac{1}{2}\log{\left(2\pi(\sigma^2 + \mbx_{N+1}^\top\mbSigma_N\mbx_{N+1})\right)} + \frac{\sigma^2 + \left(\left(\mbtheta_{\mathrm{true}} - \mbmu_N)^\top \mbx_{N+1}\right)\right)^2}{2\left(\sigma^2 + \mbx_{N+1}^\top\mbSigma_N\mbx_{N+1}\right)}
\end{align}

where we used the fact that $\E_{p(y)}[(y-\mu_{\mathrm{pred}})^2] = \V_{p(y)}[y] + (\mu_{\mathrm{true}}-\mu_{\mathrm{pred}})^2 = \sigma_{\mathrm{true}}^2 + (\mu_{\mathrm{true}}-\mu_{\mathrm{pred}})^2$ and:

\begin{align}
    \bbH[\cN(y|\mu_{\mathrm{true}},\sigma_{\mathrm{true}}^2)] &= \frac{1}{2}\log(2\pi e \sigma_{\mathrm{true}}^2)\\
    \text{KL}\left[\cN(y|\mu_{\mathrm{true}},\sigma_{\mathrm{true}}^2)||\cN(y|\mu_{\mathrm{pred}},\sigma_{\mathrm{pred}}^2)\right] &= \frac{1}{2}\log\left(\frac{\sigma_{\mathrm{pred}}^2}{\sigma_{\mathrm{true}}^2}\right) + \frac{\sigma_{\mathrm{true}}^2 + (\mu_{\mathrm{true}}-\mu_{\mathrm{pred}})^2}{2\sigma_{\mathrm{pred}}^2} - \frac{1}{2}.
\end{align}

% Finally, by replacing $\mu_{\mathrm{pred}}$ and $\sigma_{\mathrm{pred}}^2$:

% \begin{align}
%     G_N &= \E_{p(y)}[F_{N+1}] - F_N \\
%     &= -\frac{1}{2}\log{(2\pi\sigma_{\mathrm{pred}}^2)} - \frac{1}{2\sigma_{\mathrm{pred}}^2}\left(\sigma^2 + (\hat{\mbtheta}^\top \mbx_{N+1} - \mbmu_N^\top \mbx_{N+1})\right)^2 \\
%     &= -\frac{1}{2}\log{\left(2\pi(\sigma^2 + \mbx_{N+1}^\top\mbSigma_N\mbx_{N+1})\right)} - \frac{\sigma^2 + ((\hat{\mbtheta} - \mbmu_N)^\top \mbx_{N+1})^2}{2(\sigma^2 + \mbx_{N+1}^\top\mbSigma_N\mbx_{N+1})}
% \end{align}
%This can be interpreted as the negative cross-entropy between the true data generating distribution for a new point $y_{N+1}$ and the model's predictive distribution based on the previous $N$ points. 

% This quantity can also be expressed using the \gls{KL} divergence:

% \begin{align}
%     G_N &= \E_{p(y)}[\log{q_N(y)}] \\
%     &= \E_{p(y)}[\log{p(y)}] - \text{KL}[p(y)||q_N(y)] \\
%     &= -\bbH[p(y)] - \text{KL}[p(y)||q_N(y)] \\
%     &= \underbrace{-\frac{1}{2} \log{(2\pi e \sigma^2)}}_{\text{AU}(\mbx_{N+1})} - \underbrace{\text{KL}\left[\cN(y|\mu_{\mathrm{true}},\sigma_{\mathrm{true}}^2)||\cN(y|\mu_{\mathrm{pred}},\sigma_{\mathrm{pred}}^2)\right]}_{\text{EU}(\mbx_{N+1})}
% \end{align}

% where $p(y) = \cN(y|\mu_{\mathrm{true}},\sigma_{\mathrm{true}}^2)$, $q_N(y) = \cN(y|\mu_{\mathrm{pred}},\sigma_{\mathrm{pred}}^2)$ and $\bbH[p(y)] = \frac{1}{2} \log{(2\pi e \sigma^2)}$ is the differential entropy of the Gaussian distribution $p(y)$. 

This is closely related to the asymptotic expression we derived for \gls{TU} in \cref{sec:app:tu} where, if we now consider identity basis functions and a test point $\mbx_{N+1}$, it becomes:
\begin{center}
    \begin{align}
        \text{TU}(\mbx_{N+1}) &= \underbrace{\frac{1}{2}\log(2\pi e \sigma_{\mathrm{true}}^2)}_{\text{AU}(\mbx_{N+1})} + \underbrace{\frac{1}{2(N+1)}\mbx_{N+1}^\top\mbSigma_X^{-1}\mbx_{N+1}}_{\text{EU}(\mbx_{N+1})} + O\left(\frac{1}{(N+1)^2}\right)
    \end{align}
\end{center}

In terms of generalization error $G_N$, as the posterior predictive $q_N(y)$ moves closer to the true data generating distribution $p(y)$, the \gls{KL} term goes to zero and $G_N$ converges to the aleatoric component $\frac{1}{2}\log(2\pi e \sigma_{\mathrm{true}}^2)$, i.e., if $\mu_{\mathrm{pred}} = \mu_{\mathrm{true}}$ and $\sigma_{\mathrm{pred}}^2 = \sigma_{\mathrm{true}}^2$ then $\text{KL}\left[\cN(y|\mu_{\mathrm{true}},\sigma_{\mathrm{true}}^2)||\cN(y|\mu_{\mathrm{pred}},\sigma_{\mathrm{pred}}^2)\right]=0$.

\subsubsection{Asymptotic behaviors}
\label{sec:watanabe_asymptotic_expansion}
% \noteMR{This is a bit useless if left it like that, we could mention this in the Backgound section... what do you think?}

% \noteMF{Maybe it's a good idea}

In \citet{Watanabe1999, Hayashi2025} it's showed that the generalization error and the free energy have the following asymptotic expansions:

\begin{align}
    \E_n[G_N] &= \frac{\lambda}{n} - \frac{m-1}{n\log n} + o\left(\frac{1}{n\log n}\right) \\
    F_N &= nS_n + \lambda\log n - (m-1)\log \log n + O_p(1)
\end{align}

where $\lambda$ is a positive rational number, $m$ a positive integer and $\E_n[\cdot]$ is the expectation operator the overall dataset. The constant $\lambda$ is called the \gls{RLCT} or learning coefficient, since it is dominant in the leading terms of the equations above, which represents the $\E_n[G_n]-n$ and $F_n-n$ learning curves. In algebraic geometry, $m$ is called a multiplicity. In the case of regular models $\lambda = \frac{d}{2}$ and $m=1$, where $d$ is the number of model parameters.

\subsubsection{On connecting \gls{SLT} to uncertainty scaling in deep models}
Our use of Bayesian linear regression is primarily to illustrate how, in a simplified setting, generalization error can be cleanly decomposed into aleatoric and epistemic components, thereby exposing uncertainty measures that are otherwise less apparent. We are aware that this only scrapes the surface of the problem, as extending these insights to deep, over-parameterized networks remains an open challenge. Within the broader perspective of \gls{SLT} \cite{Watanabe2009}, neural networks form a singular and non-realizable class of models, for which theoretical results—such as asymptotic error bounds or explicit learning coefficients—are scarce. In particular, Bayes and Gibbs errors are governed by the \gls{RLCT}, which quantifies effective dimensionality; yet for over-parameterized architectures, the roles of \gls{RLCT} and related invariants remain poorly understood despite recent advances in local approximations, such as the \gls{LLC} \cite{Lau2024}.

\section{Uncertainty scaling: a study on linearized Laplace}
\label{sec:lla}
\subsection{Setup}
We consider a classification problem with $C$ classes on a dataset $\cD = \{(\mbx_n,y_n)\}_{n=1}^N$, where inputs $\mbx_n \in \R^D$ and targets $y_n \in \{1, \dots, C\}$. The model is a \gls{BNN} with parameters $\mbtheta \in \R^P$ mapping inputs to a vector of logits $\mbf(\mbx;\mbtheta) \in \R^C$. The likelihood is defined as $ p(\cD \mid \mbtheta) = \prod_{n=1}^N p(y_n \mid \mbf(\mbx_n;\mbtheta))$. The log-joint density (unnormalized posterior) can be written as:

\begin{equation}
    \ell(\mbtheta,\cD) = \sum_{n=1}^N \log p(y_n \mid \mbf(\mbx_n;\mbtheta)) + \log p(\mbtheta).
\end{equation}

The Laplace approximation constructs a Gaussian surrogate $q(\mbtheta) \approx p(\mbtheta \mid \cD)$ by matching the curvature of the true posterior at its peak. This is achieved via a second-order Taylor expansion of $\ell(\mbtheta,\cD)$ around the maximum a posteriori (\gls{MAP}) estimate $\mbtheta_{\text{MAP}}$:

\begin{equation}
    \ell(\mbtheta,\cD) \approx \ell(\mbtheta_{\text{MAP}},\cD) + \underbrace{\nabla \ell(\mbtheta_{\text{MAP}})^\top}_{=0} (\mbtheta - \mbtheta_{\text{MAP}}) + \frac{1}{2} (\mbtheta - \mbtheta_{\text{MAP}})^\top \nabla^2 \ell(\mbtheta_{\text{MAP}}) (\mbtheta - \mbtheta_{\text{MAP}}).
\end{equation}

Since $\mbtheta_{\text{MAP}}$ is a local maximum, the gradient term vanishes. Identifying the remaining quadratic form with the log-density of a Gaussian $\cN(\mbtheta_{\text{MAP}}, \mbSigma)$ yields the approximation parameters:

\begin{align}
    q(\mbtheta) &= \cN(\mbtheta_{\text{MAP}},\mbSigma),\\
    \mbtheta_{\text{MAP}} &= \arg\max_{\mbtheta} \ell(\mbtheta,\cD),\\
    \mbSigma &= -\left[ \nabla^2_{\mbtheta\mbtheta} \ell(\mbtheta,\cD) \Big\rvert_{\mbtheta = \mbtheta_{\text{MAP}}} \right]^{-1},
\end{align}

where the covariance $\mbSigma$ is defined as the negative inverse Hessian. 

However, computing the full Hessian is unfeasible in practice. The \gls{GGN} approximation is commonly adopted for tractability. Under the GGN, the per-data-point Hessian is approximated using the Fisher Information:

\begin{equation}
    \nabla^2_{\mbtheta\mbtheta} \log p(y \mid \mbf(\mbx;\mbtheta)) \approx -J_\mbtheta (\mbx)^\top \mbLambda(\mbf) J_\mbtheta(\mbx),
\end{equation}

where $\mbLambda(\mbf) = -\E_{y \sim p(y|\mbf)}[\nabla_{ff}^2 \log p(y|\mbf)] \in \R^{C \times C}$ is the Hessian of the negative log-likelihood w.r.t the logits (which coincides with the Fisher information metric for canonical link functions), and $J_\mbtheta (\mbx) \in \R^{C \times P}$ is the Jacobian of the outputs w.r.t the parameters.

Following \citet{Immer2021}, the \gls{GGN} approximation enables a local linearization of the network $f_\mbtheta(\mbx)$ around $\mbtheta^*$ (usually taken as $\mbtheta_{\text{MAP}}$):

\begin{equation}
    \mbf_{\text{lin}}^{\mbtheta^*} (\mbx;\mbtheta) = \mbf(\mbx;\mbtheta^*) + J_{\mbtheta^*}(\mbx)(\mbtheta - \mbtheta^*).
\end{equation}

This reduces the \gls{BNN} to a multi-output \gls{GLM}. Comparing the two predictive distributions:

\begin{align}
    p_{\text{BNN}}(y \mid \mbx,\cD) &= \E_{q(\mbtheta)} \Big[p(y \mid \mbf(\mbx;\mbtheta))\Big],\\
    p_{\text{GLM}}(y \mid \mbx,\cD) &= \E_{q(\mbtheta)} \Big[p(y \mid \mbf_{\text{lin}}^{\mbtheta^*}(\mbx;\mbtheta))\Big].
\end{align}

Notably, $p_{\text{GLM}}$ pushes the posterior Gaussian through the linearized function, resulting in a Gaussian distribution in logit space, while $p_{\text{BNN}}$ uses the original non-linear network.

\subsection{Uncertainty quantification}

To decompose the uncertainty we use the law of total variance on the one-hot encoded target variable $\mby^*$. Considering a test point $\mbx^*$, the predictive covariance (\gls{TU}) is:

\begin{equation}
    \V[\mby^* \mid \mbx^*,\cD] = \underbrace{\E_{q(\mbtheta)}\Big[\V[\mby^* \mid \mbx^*,\mbtheta]\Big]}_{\text{\gls{AU}}} + \underbrace{\V_{q(\mbtheta)}\Big[\E[\mby^* \mid \mbx^*,\mbtheta]\Big]}_{\text{\gls{EU}}},
    \label{eq:LTV}
\end{equation}

where $\E[\mby^* \mid \mbx^*, \mbtheta] = \text{softmax}(\mbf(\mbx^*; \mbtheta))$ corresponds to the model's output probabilities. Both \gls{AU} and \gls{EU} are $C \times C$ positive semi-definite matrices.

\begin{itemize}
    \item The \gls{AU} matrix captures the expected inherent noise (irreducible aleatoric uncertainty).
    \item The \gls{EU} matrix captures the model's uncertainty regarding the true class probabilities (epistemic uncertainty).
\end{itemize}

To make the computation explicit, we assume an isotropic Gaussian prior $p(\mbtheta) = \cN(\mbtheta; \mathbf{0}, \lambda^{-1}\mbI)$, implying $\nabla^2_{\mbtheta\mbtheta} \log p(\mbtheta) = -\lambda\mbI$. Consequently, the posterior covariance $\mbSigma$ is determined by the sum of the data-induced curvature (Fisher information) and the prior precision:

\begin{equation}
    \mbSigma = \left( \sum_{n=1}^N J_{\mbx_n}^\top \mbLambda(\mbf_n) J_{\mbx_n} + \lambda\mbI \right)^{-1}.
\end{equation}

We study the predictive covariance in logit space, which serves as a proxy for the \gls{EU} in probability space of \cref{eq:LTV}. Using the linearized predictive $p_{\text{GLM}}$, the covariance matrix is given by:

\begin{align}
    \mathbf{\Sigma}_{\text{logit}}(\mbx) &= \V_{q(\mbtheta)}[\mbf_{\text{lin}}^{\mbtheta^*}(\mbx;\mbtheta)] \\
    &= \V\Big[\mbf(\mbx;\mbtheta^*) + J_\mbx (\mbtheta - \mbtheta^*)\Big] \\
    &= J_\mbx \V[\mbtheta] J_\mbx^\top \\
    &= J_\mbx \mbSigma J_{\mbx}^\top \\
    &= J_\mbx \left( \sum_{n=1}^N J_{\mbx_n}^\top \mbLambda(\mbf_n) J_{\mbx_n} + \lambda\mbI \right)^{-1} J_\mbx ^\top \in \R^{C \times C}.
\end{align}

The scalar epistemic uncertainty is obtained by taking the trace of this covariance matrix (summing the variances across output dimensions):
\begin{equation}
    \text{EU}_{\text{logit}}(\mbx) = \text{Tr}\left( \mathbf{\Sigma}_{\text{logit}}(\mbx) \right).
\end{equation}

The term $J_{\mbx_n}^\top \mbLambda(\mbf_n) J_{\mbx_n}$ represents the precision contributed by the training data. If the test Jacobian $J_\mbx$ is orthogonal to the subspace spanned by the training gradients, the covariance reduces to the prior term $\lambda^{-1} J_\mbx J_\mbx^\top$.

As highlighted by \citet{Immer2021}, the Bayesian \gls{GLM} in weight space is equivalent to a \gls{GP} in function space with a particular kernel.

\subsection{Experiments: shallow models}

\begin{figure}[t]
    \centering
    \includegraphics[width=\textwidth]{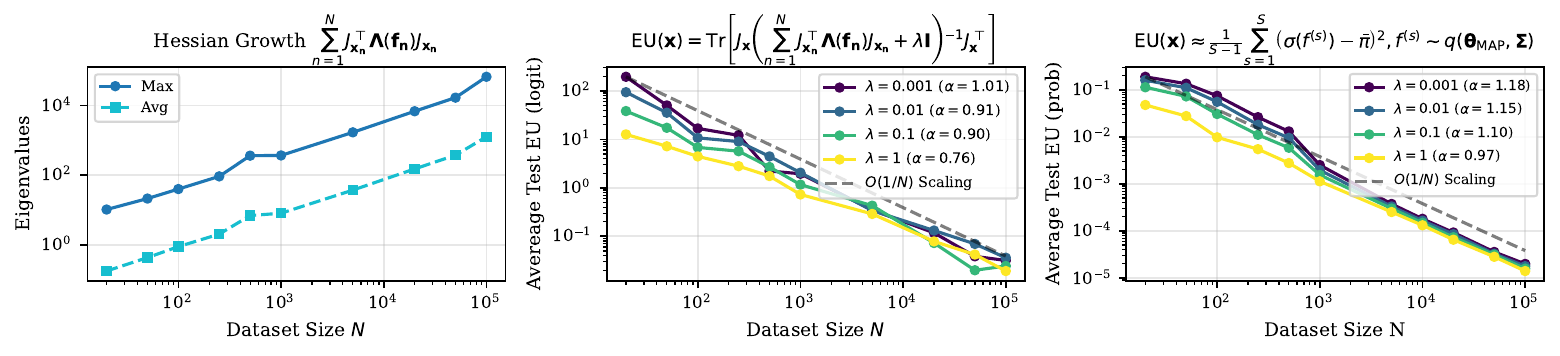}
    \caption{\textbf{Scaling Analysis of \gls{LLA}:} 
    \textbf{(Left)} Growth of the Hessian (data precision) spectrum as a function of dataset size $N$. Both the maximum eigenvalue \textcolor{blue}{(blue)} and the average eigenvalue \textcolor{cyan}{(cyan)} scale linearly with $N$, confirming that information accumulates additively.
    \textbf{(Right)} Average \gls{EU} on the test set for varying prior precisions $\lambda \in [0.001, 1]$. The dashed \textcolor{gray}{(gray)} line indicates the theoretical $O(N^{-1})$ reference scaling. The legend displays the empirical scaling exponent $\alpha$ for each prior setting (where $\text{EU} \propto N^{-\alpha}$), showing that all curves asymptotically converge to the data-driven $N^{-1}$ regime regardless of the prior strength.}
    \label{fig:lla:eu_scaling}
\end{figure}

\textbf{Dataset.} We use the \texttt{Two Moons} dataset, a binary classification problem ($C=2$) with inputs $\mbx \in \R^2$. To observe the transition from the prior-dominated to the data-dominated regime, we generated training sets of varying sizes $N \in \{5, 10, 20, 50, 100, 200, 500\}$. A fixed independent test set of $N_{\text{test}}=100$ samples was used for evaluation.

\textbf{Model Architecture.} We employed an \gls{MLP} with a single hidden layer and $\tanh$ activation functions. The default architecture consisted of $50$ hidden units. The model parameters \gls{MAP} were estimated via standard regularized cross-entropy minimization using the Adam optimizer. All experiments were repeated over 3 independent folds with different random seeds for data generation and weight initialization.

We computed the uncertainty using the Linearized Laplace approximation (GLM predictive):

\begin{enumerate}
    \item \textbf{Hessian Approximation:} We use the \gls{GGN} approximation to the Hessian and computed the \textit{full} Hessian covariance matrix $\mbSigma \in \R^{P \times P}$ (where $P$ is the number of parameters).
    
    \item \textbf{Linearization:} The predictive distribution was treated as a \gls{GLM} by linearizing the network function around the \gls{MAP} estimate.

    \item \textbf{Prior:} We train models with different prior's precision $\lambda$ to observe the shift in the "knee" of the scaling curve (the transition from prior to data dominance).
    
    \item \textbf{Uncertainty Metrics:} We quantify \gls{EU} using two complementary metrics to capture uncertainty in both function and probability space:
    
    \begin{itemize}
        \item \textbf{Variance (logit space):} For a given test point $\mbx$, we compute the trace of the predictive covariance in logit space. This serves as a proxy for the volume of the posterior in function space:
        \begin{align}
            \text{EU}_{\text{logit}}(\mbx) &= \V_{q(\mbtheta)}[\mbf_{\text{lin}}^{\mbtheta^*}(\mbx;\mbtheta)] \\
            &= \text{Tr}\left( J_\mbx \mbSigma J_\mbx^\top \right),
        \end{align}
        where $J_\mbx \in \R^{C \times P}$ is the Jacobian of the model output w.r.t parameters.

        \begin{mybox}
            \textbf{Theoretical expectations:} Assuming the training data is i.i.d., the accumulated Fisher information grows linearly with $N$. Consequently, in the data-dominated regime (large $N$), the posterior variance scales as $\mbSigma_N \propto \frac{1}{N}$. See \cref{fig:lla:eu_scaling}.
        \end{mybox}
        \item \textbf{Variance (probability space):} In the binary setting, this metric captures the variance of the predicted Bernoulli parameter $\pi = \sigma(f)$ induced by the posterior on the weights. Since the mapping from logit space to probability space is non-linear, a Gaussian distribution on $f$ does not result in a Gaussian on $\pi$. The variance is given by:
        \begin{equation}
            \text{EU}_{\text{var}}(\mbx) = \V_{q(\mbtheta)}\Big[\sigma\big(f_{\text{lin}}^{\mbtheta^*}(\mbx;\mbtheta)\big)\Big].
        \end{equation}
        Unlike the logit-space variance, this quantity is bounded in $[0, 0.25]$. Because the sigmoid saturates, high variance in logit space does not necessarily translate to high variance in probability space if the mean is far from the decision boundary. This expectation is typically intractable and is approximated via Monte Carlo sampling using $S$ samples from $q(\mbtheta)$:
        \begin{equation}
            \text{EU}_{\text{var}}(\mbx) \approx \frac{1}{S-1} \sum_{s=1}^S \left( \sigma(f^{(s)}) - \bar{\pi} \right)^2, \quad \text{where } f^{(s)} \sim q(\mbtheta_{\text{MAP}},\mbSigma).
        \end{equation}

        \begin{mybox}
            \textbf{Theoretical expectations:} Applying the Delta method (first-order Taylor expansion) to the sigmoid function, the variance in probability space is approximated by:
            \begin{equation}
                \V[\sigma(f)] \approx \left[ \frac{\partial \sigma(f)}{\partial f} \right]^2 \V[f] = \Big[ \pi(1-\pi) \Big]^2 \cdot \text{EU}_{\text{logit}}(\mbx).
            \end{equation}
            While the asymptotic rate remains $O(N^{-1})$, the magnitude is heavily dampened by the term $[\pi(1-\pi)]^2$. As the model becomes more confident (predictions $\pi$ close to $0$ or $1$), the gradient of the sigmoid vanishes, suppressing the visible \gls{EU} even if the underlying logit uncertainty is moderate.
        \end{mybox}
        \item \textbf{Entropy (probability space):} We also measure the \gls{EU} in probability space via the \gls{MI} between predictions and parameters. This is estimated using Monte Carlo sampling with $S$ samples from the \gls{GLM} predictive:
    \begin{equation}
        \text{EU}_{\text{ent}}(\mbx) = H \left[ \frac{1}{S} \sum_{s=1}^S \mbp_s \right] - \frac{1}{S} \sum_{s=1}^S H[\mbp_s],
    \end{equation}
    where $H[\cdot]$ denotes the Shannon entropy, $\mbp_s = \text{softmax}(\mbf_{\text{lin}}^{\mbtheta^*}(\mbx;\mbtheta_s))$, and $\mbtheta_s \sim \cN(\mbtheta_{\text{MAP}}, \mbSigma)$.

    \begin{mybox}
        \textbf{Theoretical expectations:} Similar to the variance metric, the \gls{MI} scales asymptotically as $O(N^{-1})$. By applying a second-order Taylor expansion to the entropy functional around the mean prediction $\bar{\mbp}$, the \gls{MI} can be approximated as:
        \begin{equation}
            \text{EU}_{\text{ent}}(\mbx) \approx \frac{1}{2} \text{Tr}\left( \mathcal{I}(\bar{\mbp}) \cdot \V[\mbp] \right),
        \end{equation}
        where $\mathcal{I}(\bar{\mbp})$ is the Fisher Information of the observation model evaluated at the mean prediction. Since the predictive variance $\V[\mbp]$ is driven by the posterior weight variance $\mbSigma \propto N^{-1}$, the entropy-based uncertainty inherits this scaling, converging linearly to zero in the large data limit.
    \end{mybox}
    \end{itemize}
\end{enumerate}

\newpage

\section{Additional results}
\label{sec:app:additional_results}

\begin{figure}[h]
  \centering
     \includegraphics[width=.8\textwidth]{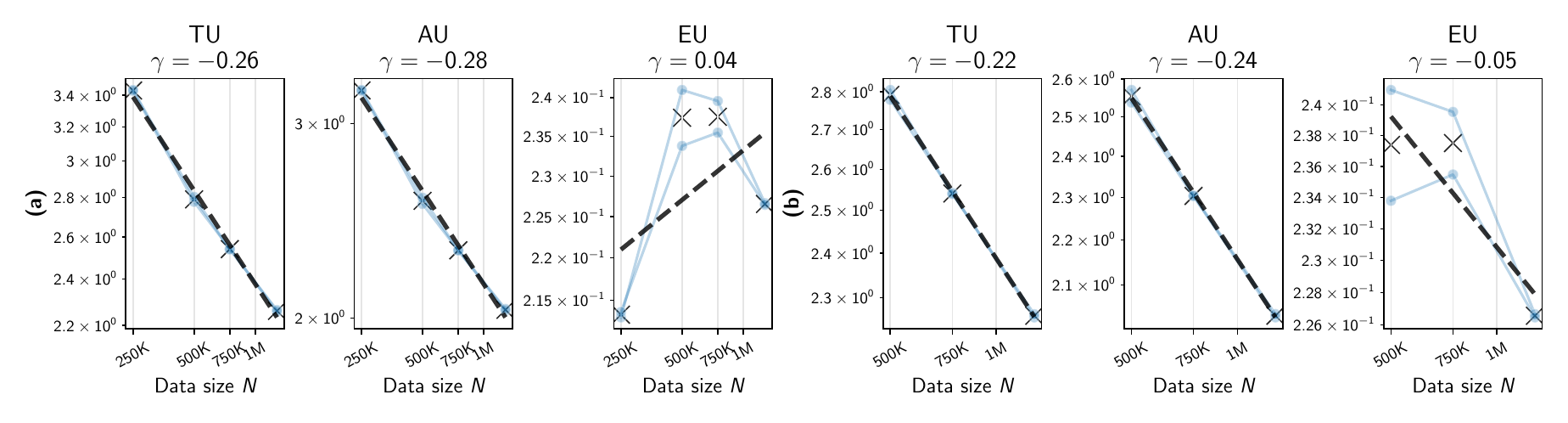}
  \caption{\textbf{Vision Transformer (ViT) on \texttt{ImageNet-32} dataset}: We use \gls{MCD} with fixed dropout rate $p=0.1$. We train for $200$ epochs using \gls{SGD} with cosine annealing. In \textbf{(a)} we report uncertainty scaling on subsets from size $250K$ up to $1.2M$. In \textbf{(b)} we discard the first point highlighting the decreasing rate of \gls{EU} for large values of $N$.}
  \label{app:fig:vit-imagenet32}
\end{figure}

% \begin{figure}[h]
%   \centering
%      \includegraphics[width=\textwidth]{figures/arxiv_figure_10_version_2.pdf}
%   \caption{\textbf{ResNets on \texttt{CIFAR-100} dataset}: We train the models with \gls{MCD} with fixed dropout rate $p=0.2$ (\textit{first row}) and $p=0.5$ (\textit{second row}). We consider $25\%$, $50\%$, $75\%$ and $100\%$ subsets of the training data. We report results from $10$ independent folds (varying both data subsampling and model initialization) showing the mean uncertainty for each $N$ subset. The dashed lines represent linear regressions fitted to the average (over the folds for every $N$) of each uncertainty metric. Both axes are on a logarithmic scale.}
%   \label{app:fig:resnets-cifar100}
% \end{figure}

\begin{figure}[h]
  \centering
     \includegraphics[width=.8\textwidth]{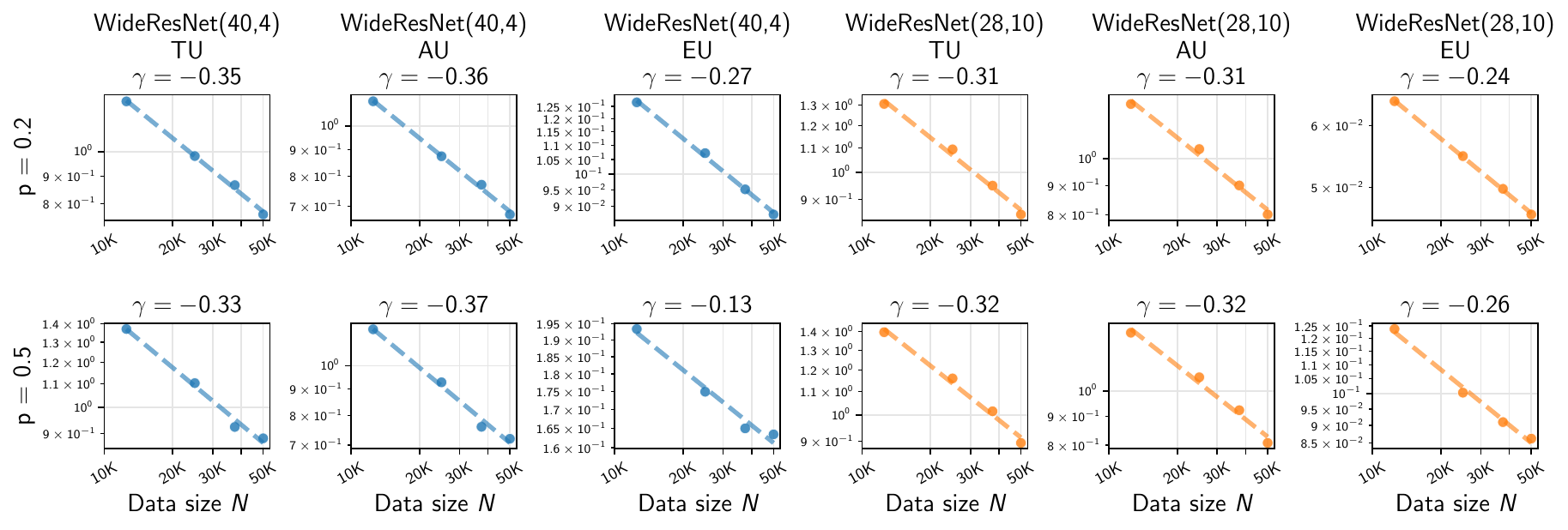}
  \caption{\textbf{WideResNets ($\mathbf{w}$,$\mathbf{d}$) on \texttt{CIFAR-100} dataset}: We train the models with \gls{MCD} with fixed dropout rate $p=0.2$ (\textit{first row}) and $p=0.5$ (\textit{second row}). We consider $25\%$, $50\%$, $75\%$ and $100\%$ subsets of the training data and we show the mean uncertainty for each $N$ subset. Dashed lines represent linear regression fitted to the test mean of each uncertainty metric. Both axes are on a logarithmic scale.}
  \label{app:fig:wideresnets-cifar100}
\end{figure}

\begin{figure}[h]
  \centering
     \includegraphics[width=0.8\textwidth]{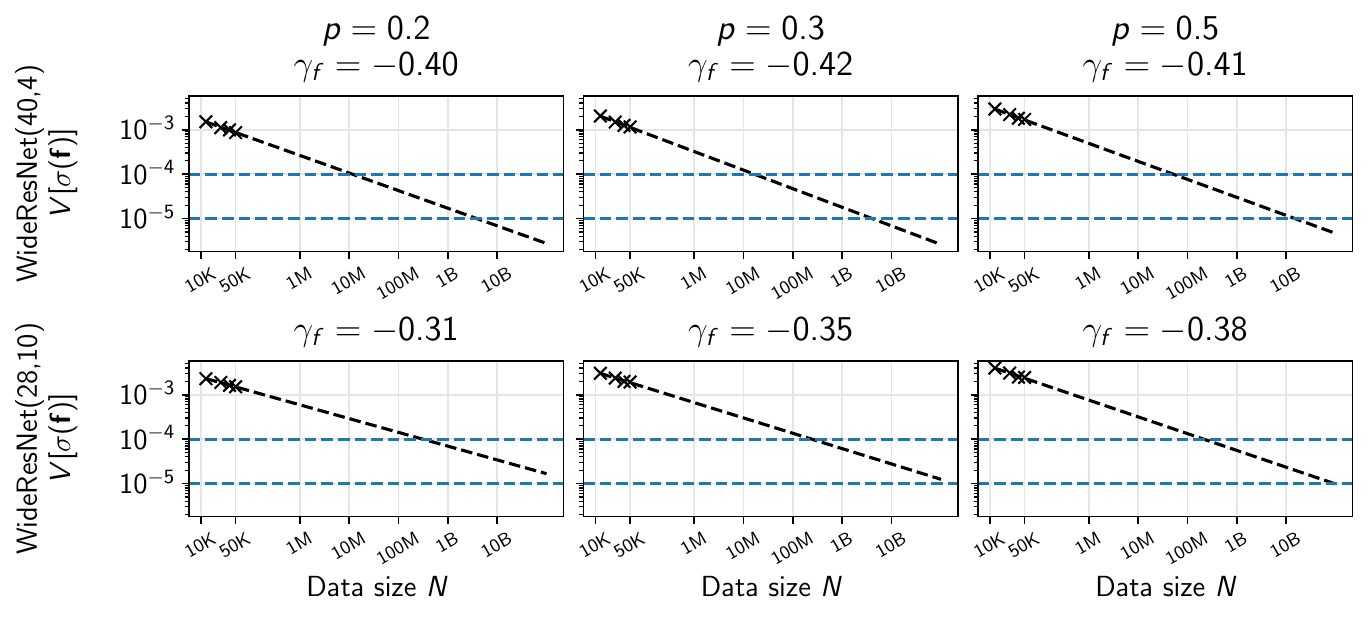}
  \caption{\textbf{WideResNets ($\mathbf{w}$,$\mathbf{d}$) on \texttt{CIFAR-10} dataset - extrapolation}: We report uncertainty trends for different WideResNets and dropout rates. We mention in the main paper that the scaling laws we derive in this context are practically useful to extrapolate uncertainties to $N$ arbitrarily large. We report $\V[\sigma(\mathbf{f})]$, the variance computed over the Softmax predictions of the \gls{MC} samples, averaged over all the test samples ($\times$ in the plots). The black dashed lines have slope $\gamma_f$, the blue dashed lines correspond to two thresholds of predictive uncertainty.}
  \label{app:fig:wideresnets-cifar10-extrapolation}
\end{figure}

\begin{figure}[h]
  \centering
     \includegraphics[width=.8\textwidth]{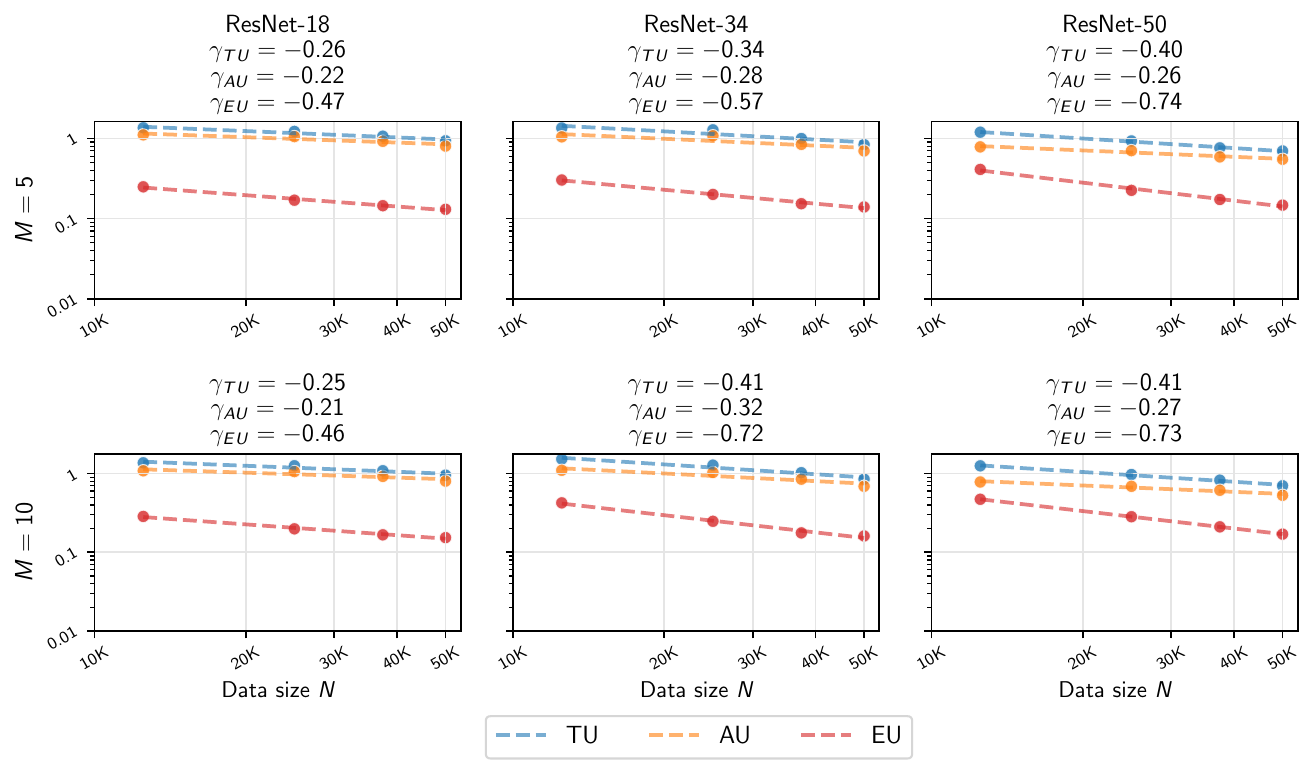}
  \caption{\textbf{ResNets on \texttt{CIFAR-100} dataset with \gls{DE}}: We train the models with \gls{DE} with ensemble members $M=5$ (\textit{first row}) and $M=10$ (\textit{second row}). We consider $25\%$, $50\%$, $75\%$ and $100\%$ subsets of the training data and we show the mean uncertainty for each $N$ subset. Dashed lines represent linear regression fitted to the test mean of each uncertainty metric. Both axes are shown on a logarithmic scale.}
  \label{app:fig:resnets-cifar100-deep_ensembles}
\end{figure}

\begin{figure}[h]
  \centering
     \includegraphics[width=.8\textwidth]{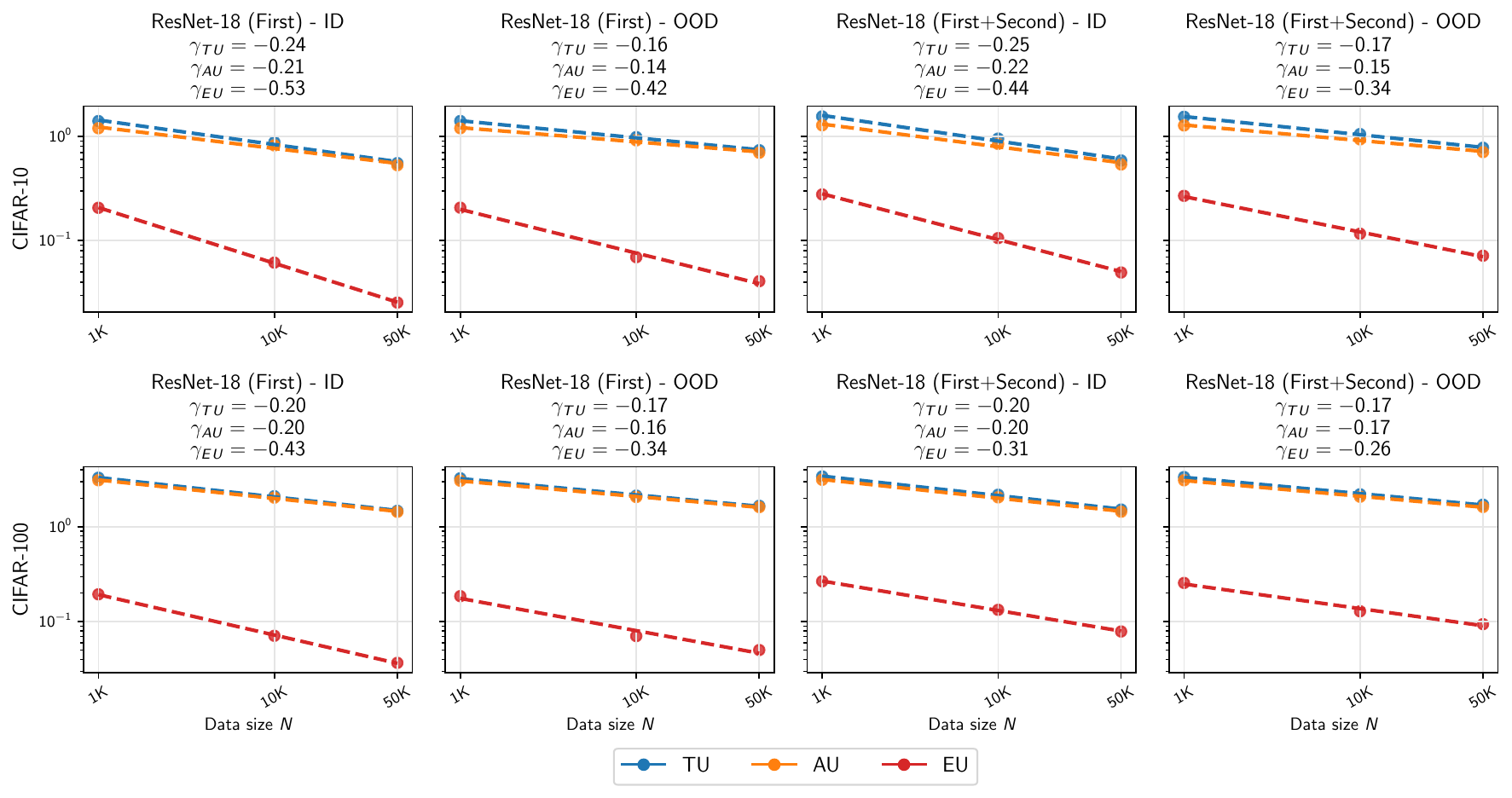}
  \caption{\textbf{ResNet-18 on \texttt{CIFAR-10} and \texttt{CIFAR-100} datasets with \gls{MCMC}}: We report uncertainties obtained through \gls{MCMC} considering only the first layer stochastic (First) and only the first two layers stochastic (First+Second). We show the scaling for both in-distribution (ID) and out-of-distribution (\gls{OOD}) predictive uncertainties. On the first row the mean uncertainty scaling for \texttt{CIFAR-10} using \texttt{CIFAR-10-C} as \gls{OOD} dataset while on the second row the mean uncertainty scaling for \texttt{CIFAR-100} using \texttt{CIFAR-100-C} as \gls{OOD} dataset.}
  \label{app:fig:resnets-cifar100-mcmc}
\end{figure}

\begin{figure}[h]
  \centering
     \includegraphics[width=.4\textwidth]{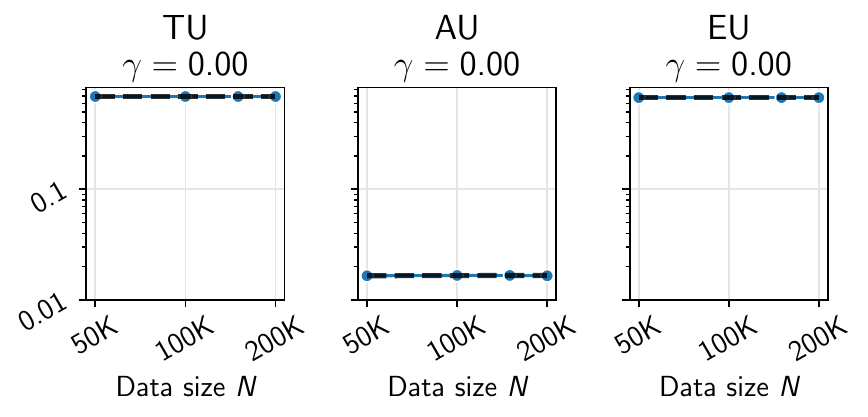}
  \caption{\textbf{Uncertainty scaling with Bayesian LoRA}: For completeness, we report the uncertainty scaling we get when fine-tuning LLM Phi on Quora Questions Pairs (\texttt{qqp}) dataset using increasing training subsets and by doing Laplace approximation over the LoRA parameters following \citet{Yang2024}. These curves are completely flat, suggesting that assessing predictive uncertainty on much smaller (fine-tuning) data compared to the massive amount used for pre-training doesn't give any insight about the scaling laws of the metrics of interest.}
  \label{app:fig:bayesian_lora}
\end{figure}

% \begin{figure}[h]
%     \centering
%     \includegraphics[width=\linewidth]{figures/MLP_experiment.pdf}
%     \caption{\textbf{Uncertainty scaling of an \gls{MLP} on the \texttt{two\_moons} dataset}: We run a toy $2D$ experiment on the \texttt{two\_moons} dataset ($10{,}000$ samples, $5{,}000$ train/test split). We use Bayesian logistic regression with normal priors on an \gls{MLP} with two hidden layers of size $32$ and $\tanh$ activation functions. Posterior inference is performed with \gls{HMC} sampling ($200$ posterior samples after $1{,}000$ warm-up steps). Training is carried out on $20$ log-spaced values of $n_{\mathrm{train}}$ and $5$ random seeds. Uncertainties are computed on the held-out $5{,}000$ test set, and we report results only for $n_{\mathrm{train}}>50$. The uncertainty is once again dominated by aleatoric contributions, while \gls{EU} clearly decays with increasing $n$. This qualitative behavior is consistently observed across all the experiments presented in this work.}
%     \label{fig:mlp_experiment}
% \end{figure}

\newpage

\ \\

\newpage

\ \\

\newpage

\ \\

\newpage

\section{Experimental setup}
\label{sec:app:experimental_setup}
\begin{table}[!h]
  \centering
  \caption{\textbf{Vision Transformer (ViT) architecture}: To obtain the results in  \cref{fig:cifar10-ViT-two-experiments-learning} and \cref{app:fig:vit-imagenet32}, we use a compact ViT model with details specified in the table below. The implementation is available at \url{https://github.com/kentaroy47/vision-transformers-cifar10}.}
  \label{tab:vit_small_arch}
  \begin{tabular}{@{}ll@{}}
    \toprule
    \textbf{Parameter}         & \textbf{Value}               \\ 
    \midrule
    Patch size                 & $4$          \\                          
    Embedding dimension        & $512$   \\
    Transformer depth          & $6$                            \\
    Number of attention heads  & $8$                            \\
    MLP hidden dimension       & $512$                          \\
    \bottomrule
  \end{tabular}
\end{table}

\begin{table}[!h]
  \centering
  \caption{\textbf{Algorithmic dataset}: To obtain the results in \cref{fig:grok} we use a GPT-2 model and train it for $10.000$ epochs with AdamW \citep{Loshchilov2019} optimizer (learning rate $10^{-4}$) and linear scheduler with $100$ warmup steps. Implementation available at \url{https://github.com/openai/grok/tree/main/scripts}.}
  \label{tab:gpt2_config}
  \begin{tabular}{@{}ll@{}}
    \toprule
    \textbf{Parameter}       & \textbf{Value}     \\          
    \midrule
    Max sequence length             & $256$                                    \\
    Embedding size                   & $128$                                    \\
    Number of layers                  & $2$                                      \\
    Number of attention heads                   & $4$        \\
    Dropout (residuals)              & $0.1$    \\
    Dropout (embeddings)             & $0.1$             \\
    Dropout (attention)              & $0.1$             \\
    \bottomrule
  \end{tabular}
\end{table}

\begin{table}[!h]
  \centering
  \caption{\textbf{\gls{IVON} configuration and training setup}: To obtain the results in \cref{fig:ivon_mcd_scale_P} we follow the setup of \citet{Shen2024} for image classification. The optimizer is combined with a scheduler (initial linear warm-up followed by cosine annealing). In this table we report the hyper-parameters we set following the same notation of \citet{Shen2024} and of the official implementation available at \url{https://github.com/team-approx-bayes/ivon}.}
  \label{tab:ivon_config}
  \begin{tabular}{@{}ll@{}}
    \toprule
    \textbf{Parameter} & \textbf{Value} \\
    \midrule
    Total epochs & $200$ \\
    Warm-up epochs & $5$ \\
    Batch size & $50$ \\
    Initial learning rate  & $2.5$ \\
    $\beta_1$  & $0.9$ \\
    $\beta_2$  & $1 - 5\times 10^{-6}$ \\
    Weight decay  & $5\times 10^{-5}$ \\
    Hessian init  & $0.05$ \\
    ESS $\lambda$  & $1{,}281{,}167$ \\
    $N$ & $1{,}281{,}167$ \\
    \gls{MC} test samples & $10$ \\
    \bottomrule
  \end{tabular}
\end{table}

% \section{\glspl{LLM} usage}

% \glspl{LLM} were employed for text polishing. They were not involved in designing experiments, analyzing data, or generating scientific content; their use was limited to improving clarity and readability of the manuscript.